%% file: main.tex
\documentclass[10pt]{article} 
\usepackage[preprint]{tmlr}

\input{math_commands.tex}

\usepackage[accsupp]{axessibility}
\usepackage{times}
\usepackage{epsfig}
\usepackage{amsmath}
\usepackage{amssymb}

\usepackage{graphicx}
\usepackage{mathtools}  
\usepackage{booktabs}       
\usepackage{amsfonts}       
\usepackage{nicefrac}       
\usepackage{caption}
\usepackage{wrapfig}
\usepackage{bm}
\usepackage{xcolor}
\usepackage{subcaption} 
\usepackage{bbm}
\usepackage{multirow}
\usepackage{makecell}

\usepackage[hidelinks]{hyperref}
\usepackage{url}
\usepackage{cleveref}
\usepackage{glossaries}

\newcommand{\defeq}{\vcentcolon=}
\DeclareMathOperator*{\expect}{\mathbb{E}}

\title{SOAP-RL: Sequential Option Advantage Propagation for Reinforcement Learning in POMDP Environments}

\author{\name Shu Ishida \email shu.ishida@oxon.org \\
      \addr Visual Geometry Group\\
      University of Oxford
      \AND
      \name Jo\~{a}o F. Henriques \email joao@robots.ox.ac.uk \\
      \addr Visual Geometry Group\\
      University of Oxford}

\def\pubmonth{07}  
\def\pubyear{2024} 
\def\openreview{\url{https://openreview.net/forum?id=XXXX}} 

\input{glossary}

\begin{document}

\maketitle

\begin{abstract}
This work compares ways of extending Reinforcement Learning algorithms to Partially Observed Markov Decision Processes (POMDPs) with \emph{options}. One view of options is as temporally extended action, which can be realized as a memory that allows the agent to retain historical information beyond the policy's context window. While option assignment could be handled using heuristics and hand-crafted objectives, learning temporally consistent options and associated sub-policies without explicit supervision is a challenge. Two algorithms, PPOEM and SOAP, are proposed and studied in depth to address this problem. 
PPOEM applies the forward-backward algorithm (for Hidden Markov Models) to optimize the expected returns for an option-augmented policy. However, this learning approach is unstable during on-policy rollouts. It is also unsuited for learning causal policies without the knowledge of future trajectories, since it optimizes option assignments for offline sequences where the entire episode is available. 
As an alternative approach, SOAP evaluates the policy gradient for an optimal option assignment. It extends the concept of the Generalized Advantage Estimation (GAE) to propagate \emph{option advantages} through time, which is an analytical equivalent to performing temporal back-propagation of option policy gradients. 
This option policy is only conditional on the history of the agent, not future actions.
Evaluated against competing baselines, SOAP exhibited the most robust performance, correctly discovering options for POMDP corridor environments, as well as on standard benchmarks including Atari and MuJoCo, outperforming PPOEM, as well as LSTM, Option-Critic, and Double Actor-Critic baselines. 
The open-sourced code is available at \textcolor{blue}{\url{https://github.com/shuishida/SoapRL}}.
\end{abstract}

\section{Introduction}
While deep \gls{rl} has seen rapid advancements in recent years, with numerous real-world applications such as robotics~\citep{gu2017deep,akkaya2019solvingrubik,haarnoja2024learning}, gaming~\citep{hasselt_double_dqn_2015,arulkumaran2019alphastar,baker2022_openai_vpt}, and autonomous vehicles~\citep{kendall2019learning,lu2022imitation}, many algorithms are limited by the amount of observation history they condition their policy on, due to the increase in computational complexity. Developing learnable embodied agents that plan over a wide spatial and temporal horizon has been a longstanding challenge in \gls{rl}. 

With a simple Markovian policy $\pi(a_t|s_t)$, the agent's ability to make decisions $a_t$ is limited by only having access to the current state $s_t$ as input. Early advances in \gls{rl} were made on tasks that either adhere to the Markov assumption that the policy and state transitions only depend on the current state, or those can be solved by frame stacking~\citep{mnih_human-level_2015} that grants the policy access to a short history. 
However, many real-world tasks are better modeled as \glspl{pomdp}~\citep{ASTROM1965174_pomdp} with a long-term temporal dependency, motivating solutions that use a bounded working memory for computational scalability. 

For \gls{pomdp} tasks, the entire history of the agent's trajectory may contain signals to inform the agent to make a more optimal decision. This is due to the reward and next state distribution $p(r_t, s_{t+1} | s_{0:t}, a_{0:t})$ being conditional on the past states and actions, not just on the current state and action.

A common approach of accommodating \glspl{pomdp} is to learn a latent representation using sequential policies, typically using a \gls{lstm}~\citep{hochreiter_lstm_1997}, \gls{gru}~\citep{cho_gru_2014} or Transformer~\citep{vaswani2017attention}. This will allow the policy to gain access to signals from the past. Differentiable planners~\citep{tamar_value_2016,lee_gated_2018,ishida2022towards} are another line of work that incorporate a learnable working memory into the system. However, these approaches have an inherent trade-off between the duration of history it can retain (defined by the policy's context window size) and the compute and training data required to learn the policy. This is because the entire history of observations within the context window have to be included in the forward pass at training time to propagate useful gradients back to the sequential policy. Another caveat is that, with larger context windows, the input space is less constrained and it becomes increasingly unlikely that the agent will revisit the same combination of states, which makes learning the policy and value function sample-expensive, and potentially unstable at inference time if the policy distribution has changed during training.

Training \gls{rl} agents to work with longer working memory is a non-trivial task, especially when the content of the memory is not pre-determined and the agent also has to learn to store information relevant to each task. 
With the tasks that the \gls{rl} algorithms are expected to handle becoming increasingly complex~\citep{dulac2021challenges,milani2023solving,chen2023end}, there is a vital need to develop algorithms that learn policies and skills that generalise to dynamic and novel environments. Many real-world tasks are performed over long time horizons, which makes it crucial that the algorithm can be efficiently trained and quickly adapted to changes in the environment.

The aim of this work is to develop an algorithm that (a) can solve problems modeled as \gls{pomdp} using memory, (b) has an input space that is tightly constrained for the policy and value function so that they are sample efficient to train, (c) only requires the current observation to be forward-passed through a neural network at a time to reduce the memory and computational requirements.

There has been considerable effort in making \gls{rl} more generalizable and efficient. Relevant research fields include \gls{hrl}~\citep{vezhnevets2017feudal,hiro2018,pateria2021hierarchical,zhang2021hierarchical}, skill learning~\citep{pertsch2020spirl,nam2022skillbased,ase_large_scale_reusable,shi2023skill}, \gls{meta-rl}~\citep{wang2016learning,duan2016rl,pearl_2019,metarl_survey} and the options framework~\citep{sutton1999between_options, Precup2000TemporalAI}, with a shared focus on learning reusable policies. In particular, this research focuses on the options framework, which extends the \gls{rl} paradigm with a \gls{hmm} that uses options to execute long-term behavior.

Options are potential keys to abstract reasoning and high-level decision-making, since they enable temporal abstraction, credit assignment, and identification of skills and subgoals. Acquiring transferable skills and composing them to execute plans, even in novel environments, are remarkable human capabilities that are instrumental in performing complex tasks with long-term objectives. Whenever one encounters a novel situation, one can still strategise by applying prior knowledge with a limited budget of additional trial and error. One way of achieving this is by abstracting away the complexity of long-term planning by delegating short-term decisions to a set of specialised low-level policies, while the high-level policy focuses on achieving the ultimate objective by orchestrating these low-level policies. 

The Option-Critic architecture~\citep{optioncritic} presents a well-formulated solution for end-to-end option discovery. The authors showed that once the option policies are learned, the Option-Critic agent can quickly adapt when the environment dynamics are changed, whereas other algorithms suffer from the changes in reward distributions.

However, there are challenges with regard to automatically learning options. A common issue is that the agent may converge to a single option that approximates the optimal policy under a Markov assumption. Additionally, learning options from scratch can be sample-inefficient due to the need to learn multiple option policies.

In the following sections, two training objectives are proposed and derived to learn an optimal option assignment. 

The first approach, \gls{ppoem}, applies \gls{em} to a \gls{hmm} describing a \gls{pomdp} for the options framework. The method is an extension of the forward-backward algorithm, also known as the Baum-Welch algorithm~\citep{baum72}, applied to options. While this approach has previously been explored~\citep{option_ml_2016,fox2017multi,Zhang2020ProvableHI,online_baum_welch_2021}, these applications were limited to 1-step \gls{td} learning. In addition, the learned options have limited expressivity due to how the option transitions are defined, and the formulation does not adequately address the problem of a scaling factor problem~\citep{bishop_ml_book}, where recursively applying the forward-backward algorithm results in exponentially diminishing magnitudes. 
In contrast, \gls{ppoem} augments the forward-backward algorithm with \gls{gae}~\citep{gae_schulman_2015}, which is a temporal generalization of \gls{td} learning, and extends the \gls{ppo}~\citep{schulman_proximal_2017} to work with options. While we showed this approach to be effective in a limited setting of a corridor environment requiring memory, the performance degraded with longer corridors. It could be hypothesized that this is due to the learning objective being misaligned with the true \gls{rl} objective, as the approach assumes access to the full trajectory of the agent for the optimal assignment of options, even though the agent only has access to its past trajectory (and not its future) at inference time. 

As an improved proposal, \gls{soap} evaluates and maximizes the policy gradient for an optimal option assignment directly. With this approach, the option policy is only conditional on the history of the agent. The derived objective has a surprising resemblance to the forward-backward algorithm, but showed more robustness when tested in longer corridor environments.
The algorithms were also evaluated on the Atari~\citep{bellemare13arcade} and MuJoCo~\citep{todorov2012mujoco} benchmarks. 
Results demonstrated that using \gls{soap} for option learning is more effective and robust than using the standard approach for learning options, proposed by the Option-Critic architecture. 

\section{Background}
\subsection{Partially Observable Markov Decision Process}
\gls{pomdp} is a special case of an \gls{mdp} where the observation available to the agent only contains partial information of the underlying state. 
In this work, $s$ is used to denote the (partial) state given to the agent, which may or may not contain the full information of the environment (which shall be distinguished from state $s$ as the underlying state~$\mathfrak{s}$).\footnote{In other literature, $o$ is used to denote the partial observation to distinguish from the underlying state $s$. While this makes the distinction explicit, many works on standard \gls{rl} algorithms assume a fully observable \gls{mdp} for their formulation, leading to conflicting notations.} This implies that the ``state'' transitions are no longer fully Markovian in a \gls{pomdp} setting, and may be correlated with past observations and actions. Hence, $p(r_t, s_{t+1} | s_{0:t}, a_{0:t})$ describes the full state and reward dynamics in the case of \glspl{pomdp}, where $s_{0:t}$ is a shorthand for $\{s_t | t_1 \leq t \leq t_2\}$, and similarly with $a_{0:t}$. 

\subsection{The Options Framework}
\label{sec:background/options_framework}

Options~\citep{sutton1999between_options, Precup2000TemporalAI} are temporally extended actions that allow the agent to make high-level decisions in the environment. Each option corresponds to a specialized low-level policy that the agent can use to achieve a specific subtask. In the Options Framework, the inter-option policy $\pi(z_t|s_t)$ and an option termination probability $\varpi(s_{t+1}, z_t)$ govern the transition of options, where $z_t$ is the current option, and are chosen from an $n$ number of discrete options $\{\mathcal{Z}_1, ..., \mathcal{Z}_n\}.$
Options are especially valuable when there are multiple stages in a task that must be taken sequentially (e.g. following a recipe) and the agent must obey different policies given similar observations, depending on the stage of the task. 

Earlier works have built upon the Options Framework by either learning optimal option selection over a pre-determined set of options~\citep{mcp_peng} or using heuristics for option segmentation~\citep{NIPS2016_f442d33f,hiro2018}, rather than a fully end-to-end approach. While effective, such approaches constrain the agent's ability to discover useful skills automatically. 
The Option-Critic architecture~\citep{optioncritic} proposed end-to-end trainable systems which learn option assignment. It formulates the problem such that inter-option policies and termination conditions are learned jointly in the process of maximizing the expected returns. 
The next section goes into detail.

\subsection{Option-Critic architecture}

As mentioned in \Cref{sec:background/options_framework}, the options framework~\citep{sutton1999between_options, Precup2000TemporalAI} formalizes the idea of temporally extended actions that allow agents to make high-level decisions. Let there be $n$ discrete options $\{\mathcal{Z}_1, ..., \mathcal{Z}_n\}$ from which $z_t$ is chosen and assigned at every time step $t$. Each option corresponds to a specialized sub-policy $\pi_\theta(a_t | s_t, z_t)$ that the agent can use to achieve a specific subtask. At $t=0$, the agent chooses an option according to its inter-option policy $\pi_\phi(z_t|s_t)$ (policy over options), then follows the option sub-policy until termination, which is dictated by the termination probability function $\varpi_\psi(s_t, z_{t-1})$. Once the option is terminated, a new option $z_t$ is sampled from the inter-option policy and the procedure is repeated.

The Option-Critic architecture~\citep{optioncritic} learns option assignments end-to-end. It formulates the problem such that the option sub-policies $\pi_\theta(a_t | s_t, z_t)$ and termination function $\varpi_\psi(s_{t+1}, z_t)$ are learned jointly in the process of maximizing the expected returns. 
The inter-option policy $\pi_\phi(z_t|s_t)$ is an $\epsilon$-greedy policy that takes an argmax $z$ of the option value function $Q_\phi(s, z)$ with $1-\epsilon$ probability, and uniformly randomly samples options with $\epsilon$ probability.
In every step of the Option-Critic algorithm, the following updates are performed for a current state $s$, option $z$, reward $r$, episode termination indicator $d \in \{0, 1\}$, next state $s'$, and discount factor $\gamma \in [0, 1)$:
\begingroup
\small
\begin{align}
\begin{split}
    Q_\phi(s, z) &\leftarrow Q_\phi(s, z) + \alpha_\phi \left\{ r + \gamma (1 - d) \left[ \left (1 - \varpi_\psi(s', z) \right) Q(s', z) + \varpi_\psi(s', z) \max_z Q_\phi(s', z) \right] - Q_\phi(s, z) \right\},\\
    \theta &\leftarrow \theta + \alpha_\theta \frac{\partial \log \pi_\theta(a | s, z)}{\partial \theta} [r + \gamma Q_\phi(s', z)],\\
    \psi &\leftarrow \psi - \alpha_\psi \frac{\partial \varpi_\psi(s', z)}{\partial \psi} [Q_\phi(s', z) - \max_z Q_\phi(s', z)].
\end{split}
\end{align}
\endgroup
Here, $\alpha_\phi$, $\alpha_\theta$ and $\alpha_\psi$ are learning rates for $Q_\phi(s,z)$, $\pi_\theta(a | s, z)$, and $\varpi_\psi(s, z)$, respectively.

\gls{ppoc}~\citep{optioncritic_ppo} builds on top of the Option-Critic architecture~\citep{optioncritic}, replacing the $\epsilon$-greedy policy over the option-values with a policy network $\pi_\varphi(z | s)$ parametrized by $\varphi$ with corresponding learning rate $\alpha_\varphi$, substituting the policy gradient algorithm with \gls{ppo}~\citep{schulman_proximal_2017} to optimize the sub-policies $\pi_\theta(a | s, z)$.

A standard \gls{ppo}'s objective is:
\begin{equation}
\label{eq:background/ppo}
\mathcal{L}_\text{PPO}(\theta) = \expect_{s, a \sim \pi} \left[\min \left(\frac{\pi_{\theta}(a|s)}{\pi_{\theta_\text{old}}(a|s)}A_t^{\text{GAE}}, \text{clip}\left(\frac{\pi_{\theta}(a|s)}{\pi_{\theta_\text{old}}(a|s)}, 1 - \epsilon, 1 + \epsilon \right) A_t^{\text{GAE}} \right)\right],
\end{equation}
where $\frac{\pi_{\theta}(a|s)}{\pi_{\theta_\text{old}}(a|s)}$ is a ratio of probabilities of taking action $a$ at state $s$ with the new policy against that with the old policy, and $A_t^{\text{GAE}}$ is the \gls{gae}~\citep{gae_schulman_2015} at time step $t$.
\gls{gae} provides a robust and low-variance estimate of the advantage function. It can be expressed as a sum of exponentially weighted multi-step \gls{td} errors:
\begin{equation}
    \label{eq:gae}
    A_t^\text{GAE} = \sum_{t'=t}^{T} (\gamma \lambda)^{t'-t} \delta_{t'},
\end{equation}
where $\delta_t = r_t + \gamma (1 - d_t) V(s_{t+1}) - V(s_t)$ is the \gls{td} error at time $t$, and $\lambda$ is a hyperparameter that controls the trade-off of bias and variance. 
Extending the definition of \gls{gae} to work with options, the update formula for \gls{ppoc} can be expressed as:

\begingroup
\small
\begin{align}
\begin{split}
    A^\text{GAE}(s, z) &\leftarrow r + \gamma V(s', z') - V(s, z) + \lambda \gamma (1-d) A^\text{GAE}(s', z'),\\
    Q_\phi(s, z) &\leftarrow Q_\phi(s, z) + \alpha_\phi A^\text{GAE}(s, z),\\    
    \theta &\leftarrow \theta + \alpha_\theta \frac{\partial \mathcal{L}_\text{PPO}(\theta)}{\partial \theta},\\
    \psi &\leftarrow \psi - \alpha_\psi \frac{\partial \varpi_\psi(s, z)}{\partial \psi} A^\text{GAE}(s, z),\\
    \varphi &\leftarrow \varphi + \alpha_\varphi \frac{\partial \log \pi_\varphi(z | s)}{\partial \varphi} A^\text{GAE}(s, z).
\end{split}
\end{align}
\endgroup

\gls{ppoc} is used as one of the baselines in this work. 

\subsection{Double Actor-Critic}
\gls{dac}~\cite{zhang2019dac} is an \gls{hrl} approach to discovering options. \gls{dac} reformulates the \gls{smdp}~\cite{sutton1999between_options} of the option framework as two hierarchical \glspl{mdp} (high-\gls{mdp} and low-\gls{mdp}). 

The low-\gls{mdp} concerns the selection of low-level actions within the currently active option.
Given the current state $s_t$, selected option $z_t$, and action $a_t$, the state and action spaces of the low-\gls{mdp} can be defined as $S_t^L = (s_t, z_t)$ and $A_t^L = a_t$. With this definition, the transition function, reward function and policy for the low-\gls{mdp} can be written as:
{
\small
\begin{align}
\begin{split}
    P_L(S_{t+1}^L | S_t^L, A_t^L) &= p((s_{t+1}, z_{t+1}) | (s_t, z_t), a_t) = P(s_{t+1} | s_t, a_t) \cdot p(z_{t+1} | s_{t+1}, z_t),\\
    R_L(S_t^L, A_t^L) &= r(s_t, a_t),\\
    \pi_L(A_t^L | S_t^L) &= \pi_L(a_t | s_t, z_t).
\end{split}
\end{align}
}

The high-\gls{mdp} handles high-level decision-making, such as selecting and terminating options. 
The state and action spaces of the high-\gls{mdp} can be defined as $S_t^H = (s_t, z_{t-1})$ and $A_t^H = z_t$. With this definition, the transition function, reward function and policy for the high-\gls{mdp} can be written as:
{
\small
\begin{align}
\begin{split}
    P_H(S_{t+1}^H | S_t^H, A_t^H) &= p((s_{t+1}, z_t) | (s_t, z_{t-1}), A_t^H) = \mathbb{I}_{A_t^H = z_t} p(s_{t+1} | s_t, z_t),\\
    R_H(S_t^H, A_t^H) &= r(s_t, z_t),\\
    \pi_H(A_t^H | S_t^H) &= \pi_H(z_t | s_t, z_{t-1}).
\end{split}
\end{align}
}

The key idea is to treat the inter-option and intra-option policies as independent actors in their respective \glspl{mdp}. Under this formulation, standard policy optimization algorithms such as \gls{ppo} can be used to optimize the policies (high-level and low-level) for both \glspl{mdp}.

\subsection{Expectation Maximization algorithm}

The \gls{em} algorithm~\citep{em_algorithm} is a well-known method for learning the assignment of latent variables, often used for unsupervised clustering and segmentation. The $k$-means clustering algorithm~\citep{kmeans_Forgy1965ClusterAO} can be considered a special case of \gls{em}. The following explanation in this section is a partial summary of Chapter 9 of~\citet{bishop_ml_book}. 

The objective of \gls{em} is to find a maximum likelihood solution for models with latent variables. Denoting the set of all observed data as $\bm{X}$, the set of all latent variables as $\bm{Z}$, and the set of all model parameters as $\Theta$, the log-likelihood function is given by:
\begin{equation}
    \log p(\bm{X} | \Theta) = \log \left\{ \sum_{\bm{Z}} p(\bm{X}, \bm{Z} | \Theta) \right\}.
\end{equation}

However, evaluating the above summation (or integral for a continuous $\bm{Z}$) over all possible latents is intractable. The \gls{em} algorithm is a way to strictly increase the likelihood function by alternating between the E-step that evaluates the expectation of a joint log-likelihood $\log p(\bm{X}, \bm{Z} | \Theta)$, and the M-step that maximizes this expectation. 

In the E-step, the current parameter estimate $\Theta_\text{old}$ (using random initialization in the first iteration, or the most recent updated parameters in subsequent iterations) is used to determine the posterior of the latents $p(\bm{Z} | \bm{X}, \Theta_\text{old})$. The joint log-likelihood is obtained under this prior. The expectation, denoted as $\mathcal{Q}(\Theta; \Theta_\text{old})$, is given by:
\begin{equation}
    \mathcal{Q}(\Theta; \Theta_\text{old}) = \expect_{\bm{Z}\sim p(\cdot | \bm{X}, \Theta_\text{old})} \left[ \log p(\bm{X}, \bm{Z} | \Theta) \right] = \sum_{\bm{Z}} p(\bm{Z} | \bm{X}, \Theta_\text{old}) \log p(\bm{X}, \bm{Z} | \Theta).
\end{equation}

In the M-step, an updated parameter estimate $\Theta_\text{new}$ is obtained by maximizing the expectation:
\begin{equation}
    \Theta_\text{new} = \argmax_\Theta \mathcal{Q}(\Theta, \Theta_\text{old}).
\end{equation}

The E-step and the M-step are performed alternately until a convergence criterion is satisfied. The \gls{em} algorithm makes obtaining a maximum likelihood solution tractable~\citep{bishop_ml_book}.

\subsection{Forward-backward algorithm}
\label{sec:ppoem/forward_backward}
\begin{figure}[h]
  \centering
  \includegraphics[width=0.6\textwidth]{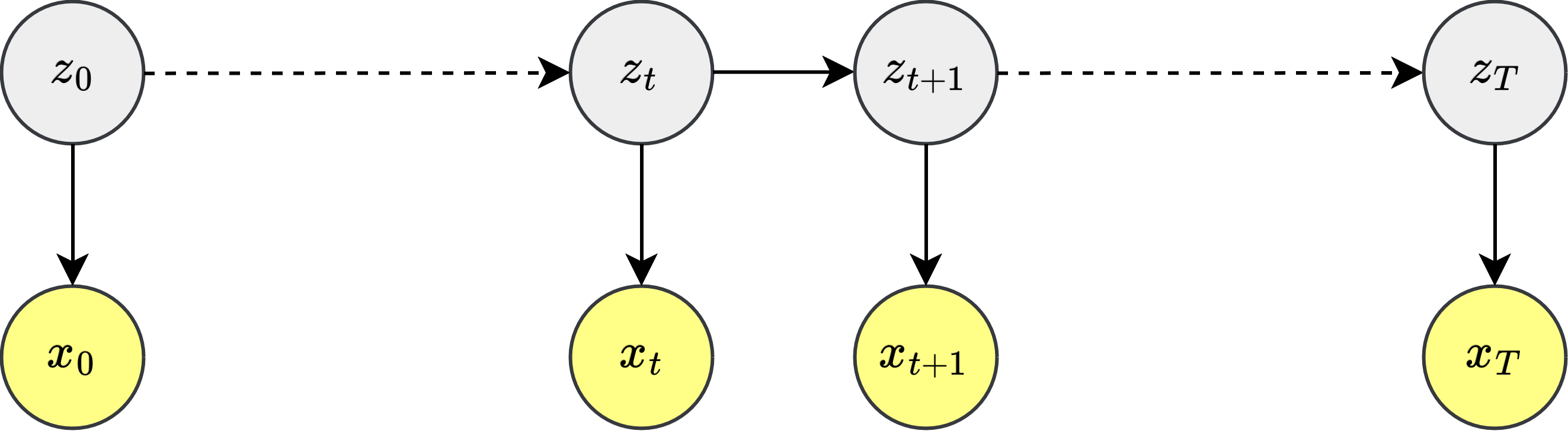}
  \caption{
    An HMM for sequential data $\bm{X}$ of length $T$, given latent variables $\bm{Z}$.
  }
  \label{fig:ppoem/hmm}
\end{figure}

The \gls{em} algorithm can also be applied in an \gls{hmm} setting for sequential data, resulting in the forward-backward algorithm, also known as the Baum-Welch algorithm~\citep{baum72}. \Cref{fig:ppoem/hmm} shows the graph of the \gls{hmm} of interest. At every time step $t \in \{0, ..., T\}$, a latent $z_t$ is chosen out of $n$ number of discrete options $\{\mathcal{Z}_1, ..., \mathcal{Z}_n\}$, which is an underlying conditioning variable for an observation $x_t$. In the following derivation, $\{x_t | t_1 \leq t \leq t_2\}$ is denoted with a shorthand $x_{t_1:t_2}$, and similarly for other variables. Chapter 13 of~\citet{bishop_ml_book} offers a comprehensive explanation for this algorithm.

For this \gls{hmm}, the joint likelihood function for the observed sequence $\bm{X} = \{x_0, ..., x_{T}\}$ and latent variables $\bm{Z} = \{z_0, ..., z_{T}\}$ is given by:
\begin{equation}
    p(\bm{X}, \bm{Z} | \Theta) = p(z_0 | \Theta) \prod_{t=0}^T p(x_t | z_t, \Theta) \prod_{t=1}^T p(z_t | z_{t-1}, \Theta).
\end{equation}

Using the above, the \gls{em} objective can be simplified as:
\begingroup
\small%
\begin{align}
\begin{split}
\label{eq:ppoem/em_objective_normal_hmm}
    \mathcal{Q}(\Theta; \Theta_\text{old}) =& \sum_{\bm{Z}} p(\bm{Z} | \bm{X}, \Theta_\text{old}) \log p(\bm{X}, \bm{Z} | \Theta)\\
    =& \sum_{z_0} p(z_0 | \Theta_\text{old}) \log p(z_0 | \Theta) + \sum_{t=0}^T \sum_{z_t} p(z_t | \bm{X}, \Theta_\text{old}) \log p(x_t | z_t, \Theta)\\
    &+ \sum_{t=1}^T \sum_{z_{t-1}, z_t} p(z_{t-1}, z_t | \bm{X}, \Theta_\text{old}) \log p(x_t, z_t | z_{t-1}, \Theta).
\end{split}
\end{align}
\endgroup

\subsection{Option policy and sub-policy}

\begin{figure}[t]
    \centering
    \begin{subfigure}{0.3\textwidth}
        \includegraphics[width=1.0\textwidth]{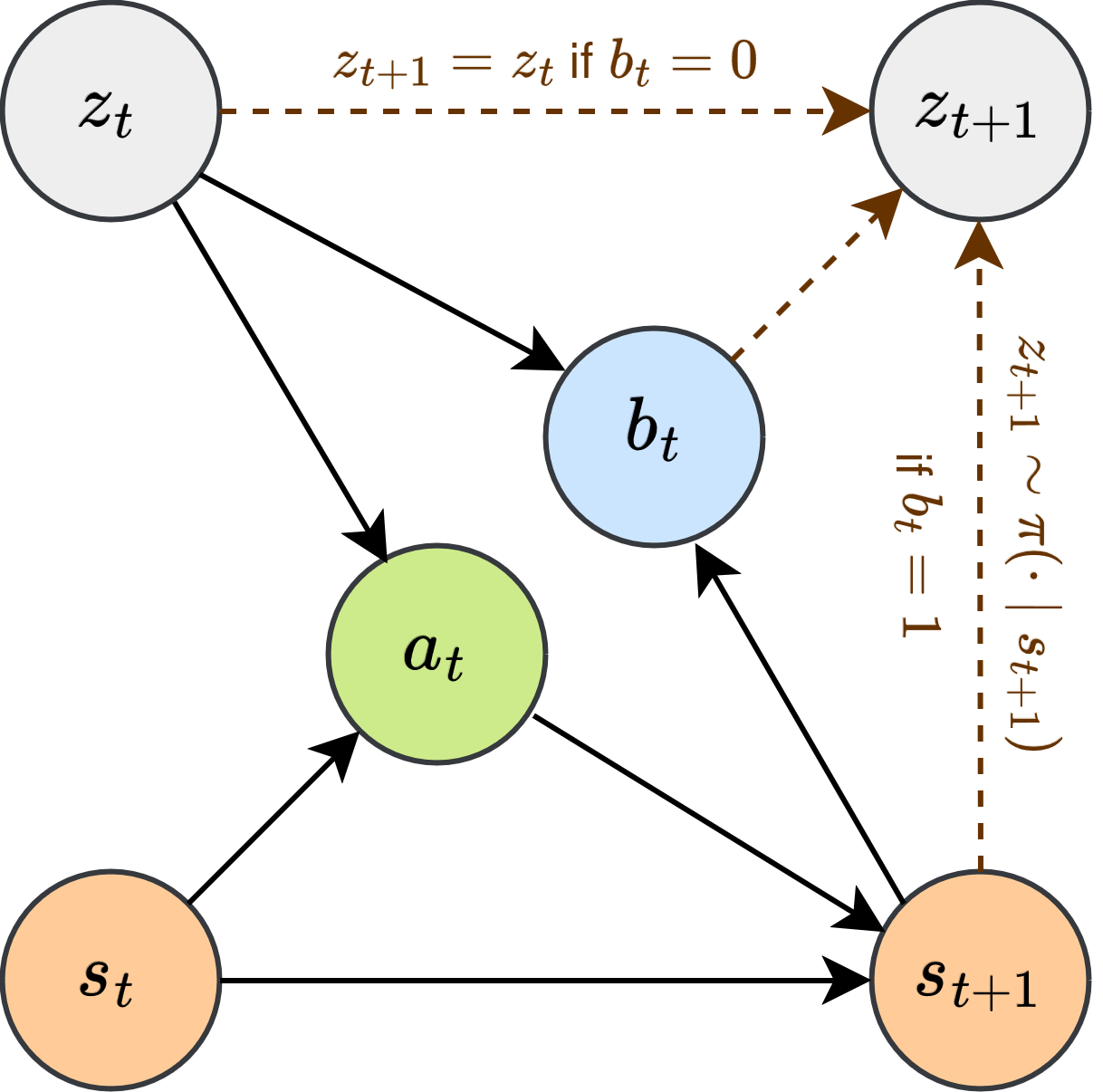}
        \caption{Standard options framework}
    \end{subfigure}
    \hspace{0.1\textwidth}
    \begin{subfigure}{0.3\textwidth}
        \includegraphics[width=1.0\textwidth]{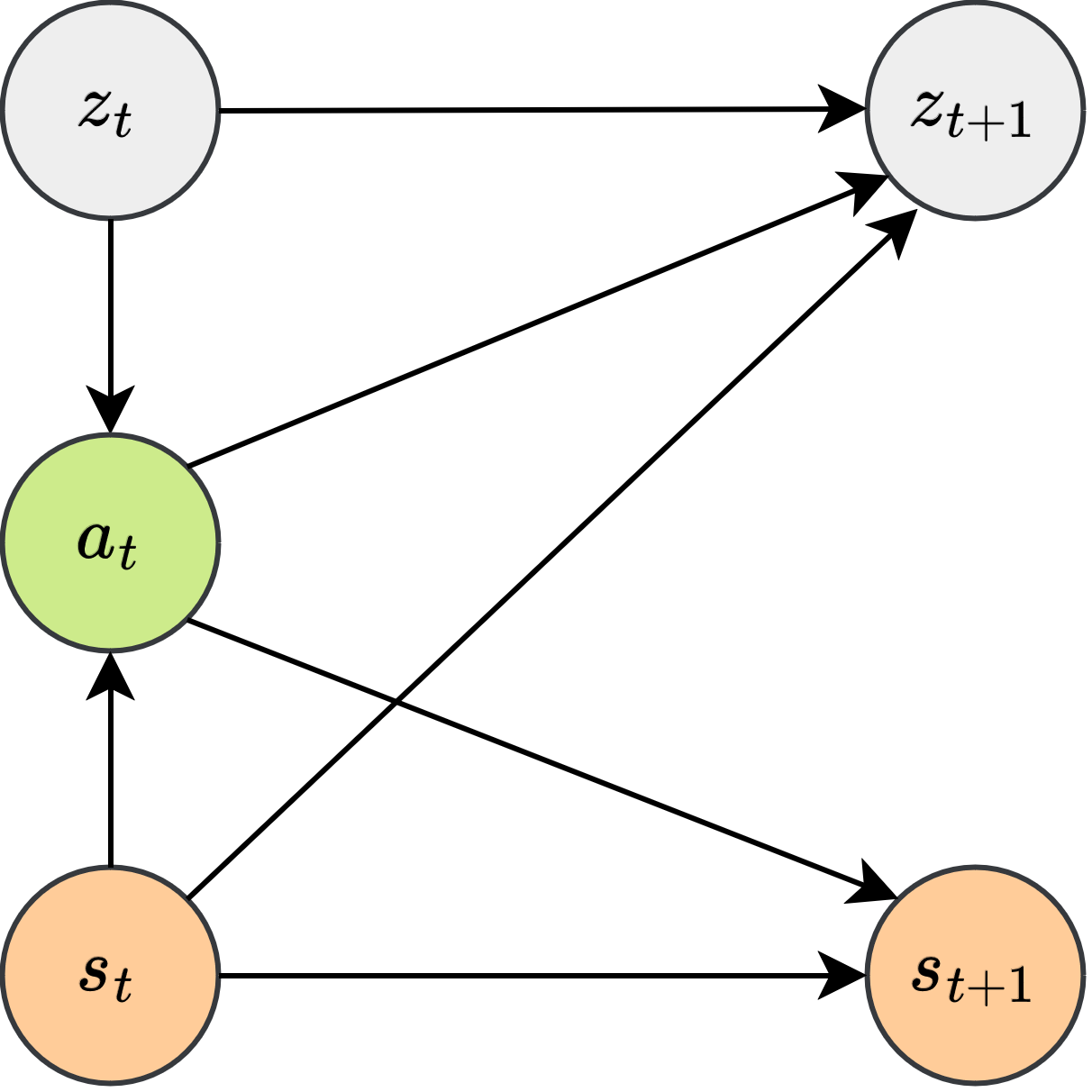}
        \caption{Options used in this work}
    \end{subfigure}
    \caption{Probabilistic graphical models showing the relationships between options $z$, actions $a$ and states $s$ at time step $t$. $b_t$ in the standard options framework denotes a boolean variable that initiates the switching of options when activated. This work adopts a more general formulation compared to the options framework, as defined in \Cref{eq:ppoem/joint_option_policy}.}
    \label{fig:ppoem/graphical_models_options}
\end{figure}

\subsubsection{E-step}
\label{sec:ppoem/e_step_normal_hmm}
In the E-step, $p(z_t | \bm{X})$ and $p(z_{t-1}, z_t | \bm{X})$ are evaluated. Note that in the following derivation, it is assumed that the probability distributions are conditioned on $\Theta$.
Defining $\alpha(z_t) \defeq p(z_t | x_{0:t})$, $\beta(z_t) \defeq \frac{p(x_{t+1:T} | z_t)}{p(x_{t+1:T} | x_{0:t})}$ and normalising constant $c_t \defeq p(x_t | x_{0:t-1})$,
\begingroup
\small%
\begin{align}
p(z_t | \bm{X}) 
&= \frac{p(x_{0:T}, z_t)}{p(x_{0:T})}= \frac{p(x_{0:t}, z_t) p(x_{t+1:T} | z_t)}{p(x_{0:t}) p(x_{t+1:T} | x_{0:t})} = \alpha(z_t)\beta(z_t),\\
p(z_{t-1}, z_t | \bm{X})
&= \frac{p(x_{0:T}, z_{t-1}, z_t)}{p(x_{0:T})}= \frac{p(x_{0:t-1}, z_{t-1}) p(x_t | z_t) p(z_t | z_{t-1}) p(x_{t+1:T} | z_t)}{p(x_{0:t-1}) p(x_t|x_{0:t-1}) p(x_{t+1:T} | x_{0:t})}\nonumber\\
&= p(x_t|z_t) p(z_t | z_{t+1}) \frac{\alpha(z_t)\beta(z_t)}{c_t}.
\end{align}
\endgroup

Recursively evaluating $\alpha(z_t)$, $\beta(z_t)$ and $c_t$,
\begingroup
\small%
\begin{align}
\alpha(z_t) &= \frac{p(x_{0:t}, z_t)}{p(x_{0:t})}
= \frac{p(x_t, z_t | x_{0:t-1})}{p(x_t | x_{0:t-1})} = \frac{\sum_{z_{t-1}} \left[ p(z_{t-1}|x_{0:t-1}) p(x_t | z_t) p(z_t | z_{t-1}) \right]}{p(x_t|x_{0:t-1})} \nonumber\\
&= \frac{p(x_t | z_t) \sum_{z_{t-1}} \left[ \alpha(z_{t-1}) p(z_t | z_{t-1}) \right]}{c_t}, \\
\beta(z_t) &= \frac{p(x_{t+1:T} | z_t)}{p(x_{t+1:T} | x_{0:t})} 
= \frac{\sum_{z_{t+1}} \left[p(x_{t+2:T} | z_{t+1}) p(x_{t+1} | z_{t+1}) p(z_{t+1} | z_t) \right]} {p(x_{t+2:T} | x_{0:t+1}) p(x_{t+1}|x_{0:t})} \nonumber\\
&= \frac{\sum_{z_{t+1}} \left[\beta(z_{t+1}) p(x_{t+1} | z_{t+1}) p(z_{t+1} | z_t) \right]} {c_{t+1}},\\
c_t &= p(x_t | x_{0:t-1}) = \sum_{z_{t-1},z_t} \left[ p(z_{t-1}|x_{0:t-1}) p(x_t | z_t) p(z_t | z_{t-1}) \right] = \sum_{z_{t-1},z_t} \left[ \alpha(z_{t-1}) p(x_t | z_t) p(z_t | z_{t-1}) \right].
\end{align}
\endgroup

The initial conditions are $\alpha(z_0) = \frac{p(x_0 | z_0)p(z_0)}{\sum_{z_0}[p(x_0 | z_0)p(z_0)]}$, $\beta(z_T) = 1$. 

\subsubsection{M-step}
In the M-step, the parameter set $\Theta$ is updated by maximizing $\mathcal{Q}(\Theta ; \Theta_\text{old})$, which can be rewritten by substituting $p(z_t | \bm{X})$ and $p(z_{t-1}, z_t | \bm{X})$ in \Cref{eq:ppoem/em_objective_normal_hmm} with $\alpha(z)$ and $\beta(z)$ (ignoring the constants) as derived in \Cref{sec:ppoem/e_step_normal_hmm}. 

\subsubsection{Option discovery via the forward-backward algorithm}
\label{sec:ppoem/background/baum_related_work}
The idea of applying the forward-backward algorithm to learn option assignments was first introduced by~\citet{option_ml_2016}, and has later been applied in both \gls{il} settings ~\citep{Zhang2020ProvableHI,online_baum_welch_2021} and \gls{rl} settings~\citep{fox2017multi}. However, in previous literature, the option policy is decoupled into an option termination probability $\varpi(s_t, z_{t-1})$, and an inter-option policy $\pi(z_t|s_t)$. Due to the inter-option policy being unconditional on the previous option $z_{t-1}$, the choice of a new option $z_t$ will be uninformed of the previous option $z_{t-1}$. This may be problematic for learning \gls{pomdp} tasks as demonstrated in \Cref{sec:ppoem/experiments/corridor}, because if a task consists of a sequence of options, then knowing which one was executed before is important to decide to move on to the next one. Previous literature also does not address the issues of exponentially diminishing magnitudes that arise from recursively applying the formula. This is known as the scaling factor problem~\citep{bishop_ml_book}. 

This work presents a concise derivation of the forward-backward algorithm applied to an improved version of the options framework. It also addresses the scaling factor problem, by building this factor into the derivation.

\section{Option assignment formulation}
The aim is to learn a diverse set of options with corresponding policy and value estimates, such that each option is responsible for accomplishing a well-defined subtask, such as reaching a certain state region. At every time step $t$, the agent chooses an option $z_t$ out of $n$ number of discrete options $\{\mathcal{Z}_1, ..., \mathcal{Z}_n\}$. 

The goal is to learn a sub-policy $\pi_{\theta}(a | s, z)$ conditional to a latent option variable $z$, and an option policy $\pi_{\psi}(z' | s, a, z)$ used to iteratively assign options at each time step, to model the joint option policy
\begin{equation}
    p_{\Theta}(a_t, z_{t+1} | s_t, z_t) = \pi_{\theta}(a_t | s_t, z_t) \pi_{\psi}(z_{t+1} | s_t, a_t, z_t).
    \label{eq:ppoem/joint_option_policy}
\end{equation} 
Here, the learnable parameter set of the policy is denoted as $\Theta = \{\theta, \psi\}$. 

\begin{figure}[t]
  \centering
  \includegraphics[width=0.9\textwidth]{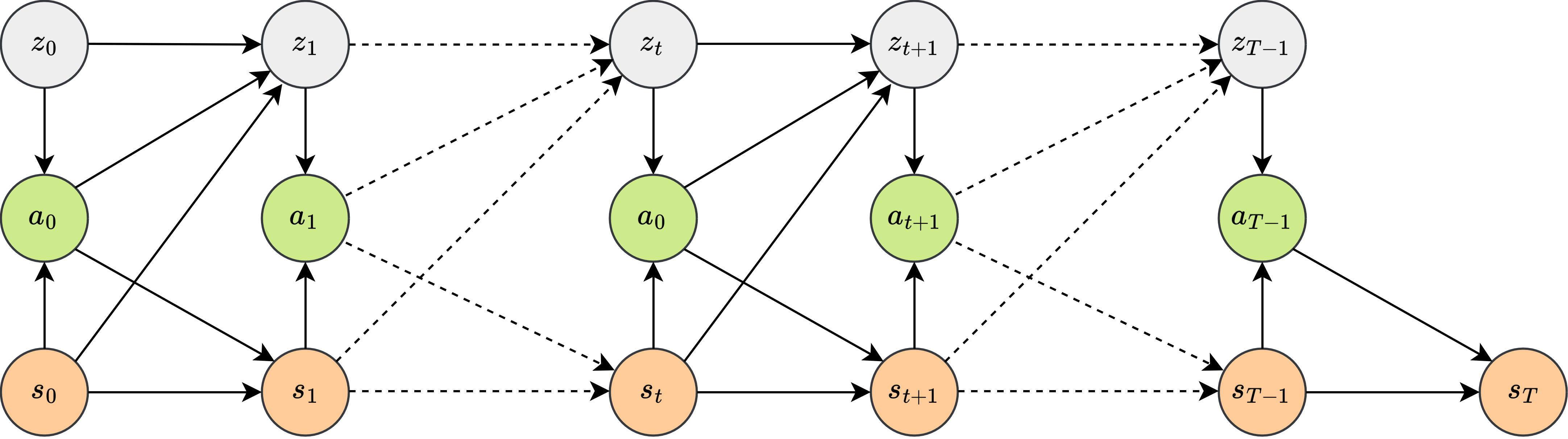}
  \caption{
    An HMM showing the relationships between options $z$, actions $a$ and states $s$. The dotted arrows indicate that the same pattern repeats where the intermediate time steps are abbreviated.
  }
  \label{fig:ppoem/full_ppoem_hmm}
\end{figure}

A comparison of the option policy used in this work and the standard options framework is shown in \Cref{fig:ppoem/graphical_models_options}. In the options framework, which further decouples the option policy $\pi_{\psi}$ into an option termination probability $\varpi(s_t, z_{t-1})$, and an unconditional inter-option policy $\pi(z_t|s_t)$. In this work, however, the option policy is modeled as $\pi_{\psi}$ with one network so that the inter-option policy is informed by the previous option $z_t$ upon choosing the next $z_{t+1}$. A graphical model for the full \gls{hmm} is shown in \Cref{fig:ppoem/full_ppoem_hmm}.

\subsection{Evaluating the probability of latents}
\label{sec:ppoem/eval_prob_latents}
Let us define an auto-regressive action probability $\alpha_t \defeq p(a_t | s_{0:t}, a_{0:t-1})$, an auto-regressive option forward distribution $\zeta(z_t) \defeq p(z_t | s_{0:t}, a_{0:t-1})$, and an option backward feedback $\beta(z_t) \defeq \frac{p(s_{t:T}, a_{t:T-1} | s_{t-1}, a_{t-1}, z_t)}{p(s_{t:T}, a_{t:T-1} | s_{0:t-1}, a_{0:t-1})}$. Notice that the definitions of action probability $\alpha$, option forward $\zeta(z_t)$, and option backward $\beta(z_t)$ resemble $c_t$, $\alpha(z_t)$ and $\beta(z_t)$ defined in \Cref{sec:ppoem/forward_backward}, respectively. 
While it is common practice to denote the forward and backward quantities as $\alpha$ and $\beta$ in the forward-backward algorithm (also known as the $\alpha$-$\beta$ algorithm), here $\alpha_t$ is redefined to denote the action probability (corresponding to the normalizing constant $c_t$), and $\zeta(z_t)$ for the option forward distribution, to draw attention to the fact that these are probabilities of option $z_t$ and action $a_t$, respectively. 

$\alpha_t$, $\zeta(z_t)$ and $\beta(z_t)$ can be recursively evaluated as follows:
\begingroup
\small%
\begin{align}
\alpha_t &= p(a_t | s_{0:t}, a_{0:t-1}) = \sum_{z_t,z_{t+1}} p(z_t | s_{0:t}, a_{0:t-1}) p_{\Theta}(a_t, z_{t+1} | s_t, z_t) = \sum_{z_t,z_{t+1}} \zeta(z_t) p_{\Theta}(a_t, z_{t+1} | s_t, z_t),\\
\zeta(z_{t+1}) &= \frac{p(z_{t+1}, s_{t+1}, a_t | s_{0:t}, a_{0:t-1})}{p(s_{t+1}, a_t | s_{0:t}, a_{0:t-1})}\nonumber\\
&= \frac{\sum_{z_t} p(z_t|s_{0:t}, a_{0:t-1}) p_{\Theta}(a_t, z_{t+1} | s_t, z_t) P(s_{t+1}|s_{0:t}, a_{0:t})}{p(a_t|s_{0:t}, a_{0:t-1}) P(s_{t+1}|s_{0:t}, a_{0:t})} \nonumber\\
&= \frac{\sum_{z_t} \zeta(z_t) p_{\Theta}(a_t, z_{t+1} | s_t, z_t)}{\alpha_t},\\
\beta(z_t) &= \frac{p(s_{t:T}, a_{t:T-1} | s_{t-1}, a_{t-1}, z_t)}{p(s_{t:T}, a_{t:T-1} | s_{0:t-1}, a_{0:t-1})} \nonumber\\
&= \frac{\sum_{z_{t+1}} \left[p(s_{t+1:T}, a_{t+1:T-1} | s_t, a_t, z_{t+1}) p_{\Theta}(a_t, z_{t+1} | s_t, z_t) P(s_t|s_{0:t-1}, a_{0:t-1}) \right]} {p(s_{t+1:T}, a_{t+1:T-1} | s_{0:t}, a_{0:t}) p(a_t|s_{0:t}, a_{0:t-1}) P(s_t|s_{0:t-1}, a_{0:t-1})} \nonumber\\
&= \frac{\sum_{z_{t+1}} \left[\beta(z_{t+1}) p_{\Theta}(a_t, z_{t+1} | s_t, z_t) \right]} {\alpha_t}.
\end{align}
\endgroup

Initial conditions are $\zeta(z_0) = p(z_0) = \frac{1}{n}$ for all possible $z_0$, indicating a uniform distribution over the options initially when no observations or actions are available, and $\beta(z_T) = \frac{p(s_T|s_{T-1}, a_{T-1}, z_{T})}{p(s_T|s_{0:T-1}, a_{0:T-1})} = 1$. 

\section{Proximal Policy Optimization via Expectation Maximization}
In this section, \gls{ppoem} is introduced, an algorithm that extends \gls{ppo} for option discovery with an \gls{em} objective. The expectation of the returns is taken over the joint probability distribution of states, actions and options, sampled by the policy. This objective gives a tractable objective to maximize, which has a close resemblance to the forward-backward algorithm.

\subsection{Expected return maximization objective with options}
The objective is to maximize the expectation of returns $R(\tau)$ for an agent policy $\pi$ over a trajectory $\tau$ with latent option $z_t$ at each time step $t$. The definition of a trajectory $\tau$ is a set of states, actions and rewards visited by the agent policy in an episode. The objective $J[\pi]$ can be written as:
\begin{align}
\begin{split}
J[\pi_\Theta] = \expect_{\tau, \bm{Z} \sim\pi}[R(\tau)] = \int_{\tau, \bm{Z}}R(\tau)p(\tau, \bm{Z} | \Theta)d\tau d\bm{Z}.
\end{split}
\end{align}

Taking the gradient of the maximization objective,
\begin{align}
\begin{split}
\label{eq:ppoem/ppoem/return_max_grad}
\nabla_\Theta J[\pi_\Theta] &= \int_{\tau, \bm{Z}}R(\tau)\nabla_\Theta p(\tau, \bm{Z} | \Theta)d\tau d\bm{Z} = \int_{\tau, \bm{Z}}R(\tau)\frac{\nabla_\Theta p(\tau, \bm{Z} | \Theta)}{p(\tau, \bm{Z} | \Theta)}p(\tau, \bm{Z} | \Theta)d\tau d\bm{Z} \\
&= \expect_{\tau, \bm{Z}}[R(\tau)\nabla_\Theta \log p(\tau, \bm{Z} | \Theta)].
\end{split}
\end{align}

To simplify the derivation, let us focus on the states and actions that appear in the trajectory. The joint likelihood function for the trajectory $\tau$ and latent options $\bm{Z} = \{z_0, ..., z_{T-1}\}$ is given by:
\begin{align}
\begin{split}
p(\tau, \bm{Z} | \Theta) &= p(s_{0:T}, a_{0:T-1}, z_{0:T} | \Theta) = p(s_0, z_0) \Pi_{t=0}^{T-1}[p_{\Theta}(a_t, z_{t+1} | s_t, z_t) P(s_{t+1} | s_{0:t}, a_{0:t})],
\end{split}
\end{align}

Evaluating $\nabla_\Theta \log p(\tau, \bm{Z} | \Theta)$, the log converts the products into sums, and the terms which are constant with respect to $\Theta$ are eliminated upon taking the gradient, leaving
\begin{align}
\begin{split}
\label{eq:ppoem/ppoem/log_joint_likelihood}
\nabla_\Theta \log p(\tau, \bm{Z} | \Theta) &= \sum_{t=0}^{T-1} \nabla_\Theta \log [\pi_{\theta}(a_t | s_t, z_t) \pi_{\psi}(z_{t+1} | s_t, a_t, z_t, s_{t+1})].
\end{split}
\end{align}

Substituting \Cref{eq:ppoem/ppoem/log_joint_likelihood} into \Cref{eq:ppoem/ppoem/return_max_grad} and explicitly evaluating the expectation over the joint option probabilities,
\begingroup
\small
\begin{align}
\begin{split}
\label{eq:ppoem/ppoem/return_max_grad_full}
\nabla_\Theta J[\pi_\Theta] &= \expect_{\tau, \bm{Z} \sim\pi} \left[\sum_{t=0}^{T-1} R(\tau) \nabla_\Theta \log p_{\Theta}(a_t, z_{t+1} | s_t, z_t) \right]\\
&= \expect_{\tau\sim\pi}\int_{\bm{Z}} \left[ \sum_{t=0}^{T-1} \left[R(\tau) \nabla_\Theta \log p_{\Theta}(a_t, z_{t+1} | s_t, z_t) \right] \right] p(\bm{Z}|\tau)\\
&= \expect_{\tau\sim\pi} \left[ \sum_{t=0}^{T-1} \sum_{z_t, z_{t+1}} \left[R(\tau) p(z_t, z_{t+1}|\tau) \nabla_\Theta \log p_{\Theta}(a_t, z_{t+1} | s_t, z_t) \right] \right].
\end{split}
\end{align}
\endgroup

Using the action probability $\alpha_t \defeq p(a_t | s_{0:t}, a_{0:t-1})$, option forward distribution $\zeta(z_t) \defeq p(z_t | s_{0:t}, a_{0:t-1})$, and option backward feedback $\beta(z_t) \defeq \frac{p(s_{t:T}, a_{t:T-1} | s_{t-1}, a_{t-1}, z_t)}{p(s_{t:T}, a_{t:T-1} | s_{0:t-1}, a_{0:t-1})}$ evaluated in \Cref{sec:ppoem/eval_prob_latents}, $p(z_t, z_{t+1} | \tau)$ can be evaluated as
\begingroup
\small%
\begin{align}
\begin{split}
\label{eq:ppoem/ppoem/joint_option_prob}
p(z_t, z_{t+1} | \tau) 
&= \frac{p(s_{0:T}, a_{0:T-1}, z_t, z_{t+1})}{p(s_{0:T}, a_{0:T-1})}\\ 
&= \frac{p(s_{0:t}, a_{0:t-1}, z_{t}) p_{\Theta}(a_t, z_{t+1} | s_t, z_t) p(s_{t+1:T}, a_{t+1:T-1} | s_t, a_t, z_{t+1})}{p(s_{0:t},a_{0:t-1}) p(a_t|s_{0:t}, a_{0:t-1}) p(s_{t+1:T}, a_{t+1:T-1} | s_{0:t}, a_{0:t})}\\ 
&= p_{\Theta}(a_t, z_{t+1} | s_t, z_t) \frac{\zeta(z_t)\beta(z_{t+1})}{\alpha_t}.
\end{split}
\end{align}
\endgroup

Using this, \Cref{eq:ppoem/ppoem/return_max_grad_full} can be evaluated and maximized with gradient descent.

\subsubsection{Relationship with Expectation Maximization}
The objective derived in \Cref{eq:ppoem/ppoem/return_max_grad_full} closely resembles the objective of the \gls{em} algorithm applied to the \gls{hmm} with options as latent variables.
The expectation of the marginal log-likelihood $\mathcal{Q}(\Theta ; \Theta_\text{old})$, which gives the lower-bound of the marginal log-likelihood $\log{p(\tau | \Theta)}$, is given by
\begin{align}
\begin{split}
\label{eq:ppoem/ppoem/em_objective}
\mathcal{Q}(\Theta; \Theta_\text{old}) 
&= \expect_{\bm{Z} \sim p(\cdot|\tau,\Theta_\text{old})}\left[\ln{p(\tau, \bm{Z} | \Theta)}\right] = \expect_{\tau\sim\pi} \int_{\bm{Z}} p(\bm{Z} | \tau, \Theta_\text{old}) \ln{p(\tau, \bm{Z} | \Theta)}d\bm{Z} \\
&= \expect_{\tau\sim\pi} \sum_{t=0}^{T-1}\sum_{z_t, z_{t+1}} \left[ p(z_t, z_{t+1} | \tau, \Theta_\text{old}) \log p_{\Theta}(a_t, z_{t+1} | s_t, z_t)\right] + \text{const.}
\end{split}
\end{align}

The difference is that the expected return maximization objective in \Cref{eq:ppoem/ppoem/return_max_grad_full} weights the log probabilities of the policy according to the returns, whereas the objective of \Cref{eq:ppoem/ppoem/em_objective} is to find a parameter set $\Theta$ that maximizes the probability that the states and actions that appeared in the trajectory are visited by the joint option policy $p_\Theta$. 

\subsection{PPO objective with Generalized Advantage Estimation}
\label{sec:ppoem/ppoem/ppoem_gae}
A standard optimization technique for neural networks using gradient descent can be applied to optimize the policy network. Noticing that the optimization objective in \Cref{eq:ppoem/ppoem/return_max_grad_full} resembles the policy gradient algorithm, the joint option policy can be optimized using the \gls{ppo} algorithm instead to prevent the updated policy $p_{\Theta}(a_t, z_{t+1} | s_t, z_t)$ from deviating from the original policy too much. 

Several changes have to be made to adapt the training objective to \gls{ppo}. Firstly, $\nabla \log p_\Theta$ is replaced by $\frac{\nabla p_\Theta}{p_{\Theta_\text{old}}}$, its first order approximation, to easily introduce clipping constraints to the policy ratios.
Secondly, the return $R(\tau)$ is replaced with the \gls{gae}, $A_t^\text{GAE}$, as described in \Cref{eq:gae}.

Extending the definition of \gls{gae} to work with options,
\begin{align}
A_t^\text{GAE}(z_t, z_{t+1} | \tau) &= r_t + \gamma V(s_{t+1}, z_{t+1}) - V(s_t, z_t) + \lambda \gamma (1-d_t) A_{t+1}^\text{GAE}(z_{t+1} | \tau)\\
A_t^\text{GAE}(z_t | \tau) &= \sum_{z_{t+1}} p(z_{t+1} | z_t, \tau) A_t^\text{GAE}(z_t, z_{t+1} | \tau).
\end{align}

The \gls{gae} could be evaluated backwards iteratively, starting from $t=T$ with the initial condition $A_T^\text{GAE}(z_{t+1} | \tau) = 0$. The option transition function $p(z_{t+1} | z_t, \tau)$ can be evaluated using $p(z_t, z_{t+1} | \tau)$ (\Cref{eq:ppoem/ppoem/joint_option_prob}) as:
\begin{equation}
    p(z_{t+1} | z_t, \tau) = \frac{p(z_t, z_{t+1} | \tau)}{\sum_{z_{t+1}} p(z_t, z_{t+1} | \tau)}.
\end{equation}

The target value $V_\text{target}(s_t, z_t)$ to regress the estimated value function towards can be defined in terms of the \gls{gae} and the current value estimate as:
\begin{equation}
V_\text{target}(s_t, z_t) = V^{\pi}(s_t, z_t) + A_t^\text{GAE}(z_t | \tau).   
\end{equation}

\section{Sequential Option Advantage Propagation}
In the previous section, assignments of the latent option variables $\bm{Z}$ were determined by maximizing the expected return for complete trajectories. The derived algorithm resembles the forward-backward algorithm closely, and requires the backward pass of $\beta(z_t)$ in order to fully evaluate the option probability $p(\bm{Z} | \tau)$. During rollouts of the agent policy, however, knowing the optimal assignment of latents $p(z_t | \tau)$ in advance is not possible, since the trajectory is incomplete and the backward pass has not been initiated. Therefore, the policy must rely on the current best estimate of the options given its available past trajectory $\{s_{0:t}, a_{0:t}\}$ during its rollout. This option distribution conditional only on its past is equivalent to the auto-regressive option forward distribution $\zeta(z_t) \defeq p(z_t | s_{0:t}, a_{0:t-1})$.

Since the optimal option assignment can only be achieved in hindsight once the trajectory is complete, this information is not helpful for the agent policy upon making its decisions. A more useful source of information for the agent, therefore, is the current best estimate of the option assignment $\zeta(z_t)$. It is sensible, therefore, to directly optimize for the expected returns evaluated over the option assignments $\zeta(z_t)$ to find an optimal option policy, rather than optimizing the expected returns for an option assignment $p(\bm{Z} | \tau)$, which can only be known in hindsight.

The following section proposes a new option optimization objective that does not involve the backward pass of the \gls{em} algorithm. Instead, the option policy gradient for an optimal forward option assignment is evaluated analytically. This results in a temporal gradient propagation, which corresponds to a backward pass, but with a slightly different outcome. Notably, this improved algorithm, \gls{soap}, applies a normalization of the option advantages in every back-propagation step through time.

As far as the authors are aware, this work is the first to derive the back-propagation of policy gradients in the context of option discovery. 

\subsection{Policy Gradient objective with options}
Let us start by deriving the policy gradient objective assuming options. The maximization objective $J[\pi]$ for the agent can be defined as:
\begin{align}
\begin{split}
J[\pi_\Theta] = \expect_{\tau \sim\pi}[R(\tau)] = \int_{\tau}R(\tau)p(\tau | \Theta) d\tau.
\end{split}
\end{align}

Taking the gradient of the maximization objective,
{\small
\begin{align}
\begin{split}
\label{eq:ppoem/soap/option_policy_grad}
\nabla_\Theta J[\pi_\Theta] &= \int_{\tau}R(\tau)\nabla_\Theta p(\tau | \Theta) d\tau = \int_{\tau}R(\tau)\frac{\nabla_\Theta p(\tau | \Theta)}{p(\tau | \Theta)}p(\tau | \Theta) d\tau = \expect_{\tau}[R(\tau)\nabla_\Theta \log p(\tau | \Theta)].
\end{split}
\end{align}
}

So far, the above derivation is the same as the normal policy gradient objective without options. Note, however, that this is NOT identical to \Cref{eq:ppoem/ppoem/return_max_grad}. Next, the likelihood for the trajectory $\tau$ is given by:
\begin{align}
\begin{split}
p(\tau | \Theta) &= p(s_{0:T}, a_{0:T-1} | \Theta) = \rho(s_0) \Pi_{t=0}^{T-1}[p(a_t | s_{0:t}, a_{0:t-1}, \Theta) P(s_{t+1} | s_{0:t}, a_{0:t})].
\end{split}
\end{align}

This is where options become relevant, as the standard formulation assumes that the policy $\pi(a | s)$ is only dependent on the current state without history, and similarly that the state transition environment dynamics $P(s' | s, a)$ is Markovian given the current state and action. In many applications, however, the \emph{states} that are observed do not contain the entire information about the underlying dynamics of the environment\footnote{Some literature on \gls{pomdp} choose to make this explicit by denoting the partial observation available to the agent as observation $o$, distinguishing from the underlying ground truth state. However, since $o$ can also stand for \emph{options}, and is used in other literature on options, here the input to the agent's policy and value functions is denoted using the conventional $s$ to prevent confusion.}, and therefore, conditioning on the history yields a different distribution of future states compared to conditioning on just the current state. To capture this, the policy and state transitions are now denoted to be $p(a_t | s_{0:t}, a_{0:t-1})$ and $P(s_{t+1} | s_{0:t}, a_{0:t})$, respectively. Here, the probabilities are conditional on the historical observations ($s_{0:t}$) and historical actions (e.g. $a_{0:t}$), rather than just the immediate state $s_t$ and action $a_t$. Note that $p(a_t | s_{0:t}, a_{0:t-1})$ is a quantity $\alpha_t$ that has already been evaluated in \Cref{sec:ppoem/eval_prob_latents}.

Evaluating $\nabla_\Theta \log p(\tau | \Theta)$, the logarithm converts the products into sums, and the terms that are constant with respect to $\Theta$ are eliminated upon taking the gradient, leaving
\begin{align}
\begin{split}
\label{eq:ppoem/soap/log_likelihood_pomdp}
\nabla_\Theta \log p(\tau | \Theta) &= \sum_{t=0}^{T-1} \nabla_\Theta \log p(a_t | s_{0:t}, a_{0:t-1}, \Theta) = \sum_{t=0}^{T-1} \nabla_\Theta \log \alpha_t = \sum_{t=0}^{T-1} \frac{\nabla_\Theta \alpha_t}{\alpha_t},
\end{split}
\end{align}
where $\alpha_t$ is substituted following its definition in \Cref{sec:ppoem/eval_prob_latents}.

Substituting \Cref{eq:ppoem/soap/log_likelihood_pomdp} into \Cref{eq:ppoem/soap/option_policy_grad},
\begin{align}
\begin{split}
\label{eq:ppoem/soap/option_policy_grad_simple}
\nabla_\Theta J[\pi_\Theta] &= 
\expect_{\tau}[R(\tau)\nabla_\Theta \log p(\tau | \Theta)] = 
\expect_{\tau \sim\pi} \left[\sum_{t=0}^{T-1} R(\tau) \frac{\nabla_\Theta \alpha_t}{\alpha_t} \right].
\end{split}
\end{align}

Similarly to \Cref{sec:ppoem/ppoem/ppoem_gae}, it is possible to substitute the return $R(\tau)$ with \gls{gae}, thereby reducing the variance in the return estimate. 
Extending the definition of \gls{gae} to work with options,
\begingroup
\small%
\begin{align}
A_t^\text{GAE}(z_t, z_{t+1}) &= r_t + \gamma V(s_{t+1}, z_{t+1}) - V(s_t, z_t) + \lambda \gamma (1-d_t) A_{t+1}^\text{GAE}(z_{t+1}),\\
A_t^\text{GAE}(z_t) &= \sum_{z_{t+1}} p(z_{t+1} | s_t, a_t, z_t) A_t^\text{GAE}(z_t, z_{t+1}),\\
V_\text{target}(s_t, z_t) &= V^{\pi}(s_t, z_t) + A_t^\text{GAE}(z_t)\label{eq:ppoem/soap/value}.
\end{align}
\endgroup

Notice that, while the definition of these estimates is almost identical to \Cref{sec:ppoem/ppoem/ppoem_gae}, the advantages are now propagated backwards via the option transition $p(z_{t+1} | s_t, a_t, z_t)$ rather than $p(z_{t+1} | z_t, \tau)$.

Substituting the \gls{gae} into \Cref{eq:ppoem/soap/option_policy_grad_simple},
\begingroup
\small%
\begin{align}
\begin{split}
\label{eq:ppoem/soap/policy_gradient_objective}
&\nabla_\Theta J[\pi_\Theta] = 
\expect_{\tau \sim\pi}\left[\sum_{t=0}^{T-1} \frac{\sum_{z_t} A_t^\text{GAE}(z_t) \zeta(z_t)}{\alpha_t} \nabla_\Theta \alpha_t \right]\\
&= \expect_{\tau \sim\pi}\left[ \sum_{t=0}^{T-1} \frac{\sum_{z_t} A_t^\text{GAE}(z_t) \zeta(z_t)}{\alpha_t} \sum_{z_t,z_{t+1}} \left[ p_{\Theta}(a_t, z_{t+1} | s_t, z_t) \nabla \zeta(z_t) + \zeta(z_t) \nabla p_{\Theta}(a_t, z_{t+1} | s_t, z_t) \right] \right].
\end{split}
\end{align}
\endgroup

\subsection{Analytic back-propagation of the policy gradient}
If a forward pass of the policy can be made in one step over the entire trajectory, a gradient optimization on the objective can be performed directly. However, this would require storing the entire trajectory in \gls{gpu} memory, which is highly computationally intensive. Instead, this section analytically evaluates the back-propagation of gradients of the objective so that the model can be trained on single time-step rollout samples during training.

Gradient terms appearing in \Cref{eq:ppoem/soap/policy_gradient_objective} are either $\nabla \zeta(z_t)$ or $\nabla p_{\Theta}(a_t, z_{t+1} | s_t, z_t)$ for $0 \leq t \leq T-1$. 
While $p_{\Theta}(a_t, z_{t+1} | s_t, z_t)$ is approximated by neural networks and can be differentiated directly, $\nabla \zeta(z_{t+1})$ has to be further expanded to evaluate the gradient in recursive form as:
\begingroup
\small%
\begin{align}
\begin{split}
\label{eq:ppoem/soap/zeta_grad}
\nabla \zeta(z_{t+1}) &= \frac{\nabla \sum_{z_t} \zeta(z_t) p_{\Theta}(a_t, z_{t+1} | s_t, z_t)}{\alpha_t} - \zeta(z_{t+1}) \frac{\nabla \alpha_t}{\alpha_t}\\
&= \frac{1}{\alpha_t} \left[ \sum_{z_t} \nabla \left[ \zeta(z_t) p_{\Theta}(a_t, z_{t+1} | s_t, z_t)\right] - \zeta(z_{t+1}) \sum_{z'_t, z'_{t+1}} \nabla \left[ \zeta(z'_t) p_{\Theta}(a_t, z'_{t+1} | s_t, z'_t)\right] \right].
\end{split}
\end{align}
\endgroup

Using \Cref{eq:ppoem/soap/zeta_grad}, it is possible to rewrite the $\nabla \zeta(z_{t+1})$ terms appearing in \Cref{eq:ppoem/soap/policy_gradient_objective} in terms of $\nabla \zeta(z_t)$ and $\nabla p_{\Theta}(a_t, z_{t+1} | s_t, z_t)$. Defining the coefficients of $\nabla \zeta(z_{t+1})$ in \Cref{eq:ppoem/soap/policy_gradient_objective} as option utility $U(z_{t+1})$,

\begingroup
\small%
\begin{align}
\begin{split}
&\sum_{z_{t+1}} U(z_{t+1}) \nabla \zeta(z_{t+1}) 
= \frac{1}{\alpha_t} \sum_{z_t, z_{t+1}} \left[ U(z_{t+1}) - \sum_{z'_{t+1}} U(z'_{t+1}) \zeta(z'_{t+1})  \right] \nabla \left[ \zeta(z_t) p_{\Theta}(a_t, z_{t+1} | s_t, z_t)\right]\\
&= \frac{1}{\alpha_t} \sum_{z_t, z_{t+1}} \left[ U(z_{t+1}) - \sum_{z'_{t+1}} U(z'_{t+1}) \zeta(z'_{t+1})  \right] \left[ p_{\Theta}(a_t, z_{t+1} | s_t, z_t) \nabla \zeta(z_t) +  \zeta(z_t) \nabla p_{\Theta}(a_t, z_{t+1} | s_t, z_t) \right].
\end{split}
\end{align}
\endgroup

Thus, the occurrences of gradients $\nabla \zeta(z_{t+1})$ have been reduced to terms with $\nabla \zeta(z_t)$ and $\nabla p_{\Theta}(a_t, z_{t+1} | s_t, z_t)$.

Applying this iteratively to \Cref{eq:ppoem/soap/policy_gradient_objective}, starting with $t=T-1$ in reverse order, \Cref{eq:ppoem/soap/policy_gradient_objective} could be expressed solely in terms of gradients $\nabla p_{\Theta}(a_t, z_{t+1} | s_t, z_t)$. Defining the coefficients of $\nabla p_{\Theta}(a_t, z_{t+1} | s_t, z_t)$ as policy gradient weighting $W_t(z_t, z_{t+1})$, 
\begingroup
\small%
\begin{align}
\begin{split}
A_t^\text{GOA}(z_{t+1}) &= \sum_{z_t} A_t^\text{GAE}(z_t) \zeta(z_t) + (1 - d_t) \left[ U(z_{t+1}) - \sum_{z'_{t+1}} U(z'_{t+1}) \zeta(z'_{t+1})  \right],\\
U(z_t) &= \frac{\sum_{z_{t+1}} A_t^\text{GOA}(z_{t+1}) p_{\Theta}(a_t, z_{t+1} | s_t, z_t)}{\alpha_t},\\
W(z_t, z_{t+1}) &= \frac{A_t^\text{GOA}(z_{t+1})\zeta(z_t)}{\alpha_t}.
\end{split}
\end{align}
\endgroup
where $A_t^\text{GOA}(z_{t+1})$ is a new quantity derived and introduced in this work as \gls{goa}, which is a term that appears in evaluating $U(z_t)$ and $W(z_t, z_{t+1})$.

Rewriting the policy gradient objective in \Cref{eq:ppoem/soap/policy_gradient_objective} with the policy gradient weighting,
\begin{align}
\begin{split}
\label{eq:ppoem/soap/soap_policy_grad}
\nabla_\Theta J[\pi_\Theta] 
&= \expect_{\tau \sim\pi} \left[\sum_{t=0}^{T-1} \sum_{z_t, z_{t+1}} \frac{A_t^\text{GOA}(z_{t+1})\zeta(z_t)}{\alpha_t} \nabla_\Theta p_{\Theta}(a_t, z_{t+1} | s_t, z_t) \right].
\end{split}
\end{align}

\subsection{Learning objective for option-specific policies and values}
The training objective given in \Cref{eq:ppoem/soap/soap_policy_grad} is modified so that it could be optimized with \gls{ppo}. Unlike in \Cref{sec:ppoem/ppoem/ppoem_gae}, the training objective is written in terms of $\nabla p_\Theta$ and not $\nabla \log p_\Theta$. Therefore, the clipping constraints are applied to $p_\Theta$ directly, limiting it to the range of $(1-\epsilon)p_{\Theta_\text{old}}$ and $(1+\epsilon)p_{\Theta_\text{old}}$. The resulting \gls{ppo} objective is:
\begingroup
\small%
\begin{align}
\begin{split}
J_\Theta =& \expect_{s_t, a_t\sim\pi} \sum_{z_t, z_{t+1}} \biggl[\frac{\zeta(z_t)}{\alpha_t} \min \Bigl( \pi_{\Theta} A_t^{\text{GOA}}(z_{t+1}),\text{clip} \Bigl(\pi_{\Theta}, (1 - \epsilon) \pi_{\Theta_\text{old}}, (1 + \epsilon) \pi_{\Theta_\text{old}} \Bigr)A_t^{\text{GOA}}(z_{t+1}) \Bigr) \biggr],\\
&\textrm{with}\;\pi_{\Theta}(a_t, z_{t+1}|s_t, z_t)\textrm{ and }\pi_{\Theta_\text{old}}(a_t, z_{t+1}|s_t, z_t).
\end{split}
\end{align}
\endgroup

The option-specific value function $V_\phi^\pi(s_t, z_t)$ parameterized by $\phi$ can be learned by regressing towards the target values $V_\text{target}(s_t, z_t)$ evaluated in \Cref{eq:ppoem/soap/value} for each state $s_t$ and option $z_t$ sampled from the policy and option-forward probability, respectively. Defining the objective function for the value regression as $J_\phi$,
\begingroup
\small%
\begin{equation}
J_\phi = - \expect_{s_t\sim\pi, z_t \sim \zeta} \left[ V_\text{target}(s_t, z_t) - V_\phi^{\pi}(s_t, z_t)\right]^2.
\end{equation}
\endgroup

The final training objective is to maximize the following sum of the policy loss and value loss:
\begin{align}
\begin{split}
J_\text{SOAP} = J_\Theta + J_\phi.
\end{split}
\end{align}

\section{Experiments}

Experiments were conducted on a variety of \gls{rl} agents.
\gls{ppo}~\citep{schulman_proximal_2017} is a baseline without memory, \gls{ppoc}~\citep{optioncritic_ppo} implements the Option-Critic algorithm using \gls{ppo} for policy optimization, \gls{ppo_lstm} implements a recurrent policy with latent states using an \gls{lstm}, \gls{dac}~\citep{zhang2019dac} optimizes both the inter- and intra-option policies on hierarchical \glspl{mdp}, \gls{ppoem} is the algorithm developed in the first half of this chapter that optimizes the expected returns using the forward-backward algorithm, and \gls{soap} is the final algorithm proposed in this chapter that uses an option advantage derived by analytically evaluating the temporal propagation of the option policy gradients. \gls{soap} mitigates the deficiency of \gls{ppoem} that the training objective optimizes the option assignments over a full trajectory which is typically only available in hindsight; \gls{soap} optimizes the option assignments given only the history of the trajectory instead, making the optimization objective better aligned with the task objective.

The aim is to (a) show and compare the option learning capability of the newly developed algorithms, and (b) assess the stability of the algorithms on standard \gls{rl} environments.
All algorithms use \gls{ppo} as the base policy optimizer, and share the same backbone and hyperparameters, making it a fair comparison. All algorithms use Stable Baselines 3~\citep{stable_baselines3} as a base implementation with the recommended tuned hyperparameters for each environment. In the following experiments, the number of options was set to $4$. 

\subsection{Option learning in corridor environments}
\label{sec:ppoem/experiments/corridor}
A simple environment of a corridor with a fork at the end is designed as a minimalistic and concrete example where making effective use of latent variables to retain information over a sequence is necessary to achieve the agent's goal. 

\begin{figure}[h]
  \centering
  \includegraphics[width=1.0\textwidth]{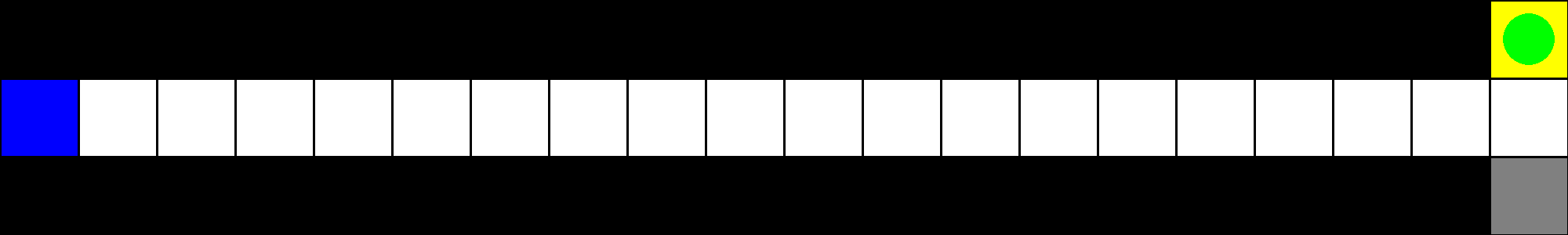}\\
  \vspace{0.5em}
  \includegraphics[width=1.0\textwidth]{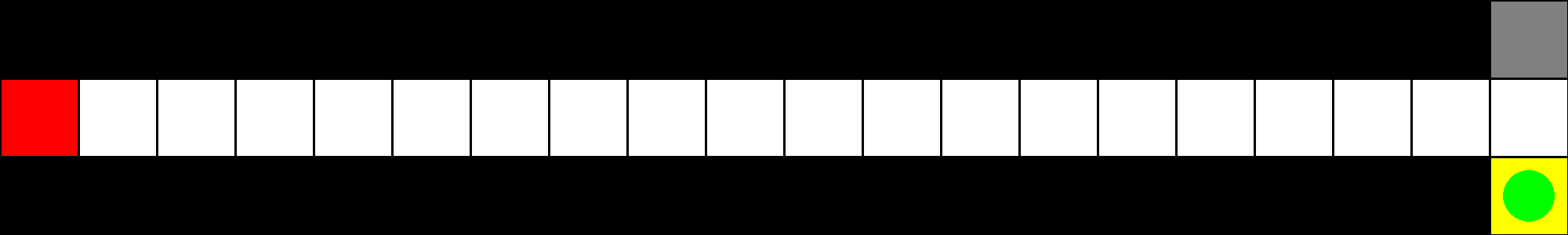}
  \caption{
    A corridor environment. The above example has a length $L=20$. The agent represented as a green circle starts at the left end of the corridor, and moves towards the right. When it reaches the right end, the agent can either take an up action or a down action. This will either take the agent to a yellow cell or a grey cell. The yellow cell gives a reward of $1$, while the grey cell gives a reward of $-1$. All other cells give a reward of $0$. The location of a rewarding yellow cell and the penalizing grey cell are determined by the color of the starting cell (either ''blue'' or ''red''), as shown, and this is randomized, each with $50\%$ probability. The agent only has access to the color of the current cell as observation. For simplicity of implementation, the agent's action space is \{''up'', ''down''\}, and apart from the fork at the right end, taking either of the actions at each time step will move the agent one cell to the right.
    The images shown are taken from rollouts of the SOAP agent after training for $100k$ steps. The agent successfully navigated to the rewarding cell in both cases.
  }
  \label{fig:ppoem/experiments/corridor}
\end{figure}

\Cref{fig:ppoem/experiments/corridor} describes the corridor environment, in which the agent has to determine whether the rewarding cell (colored yellow) is at the top or bottom, based on the color of the cell it has seen at the start (either ''blue'' or ''red''). However, the agent only has access to the color of the current cell, and does not have a bird's-eye-view of the environment. Hence, the agent must retain the information of the color of the starting cell in memory, whilst discarding all other information irrelevant to the completion of the task. The agent must learn that the information of the color of the starting cell is important to task completion in an unsupervised way, just from the reward signals. This makes the task challenging, as only in hindsight (after reaching the far end of the corridor) is it clear that this information is useful to retain in memory, but if this information was not written in memory in the first place then credit assignment becomes infeasible. The learned options can be interpreted as ``move up at the end of the corridor'' and ``move down at the end''.

\begin{figure}[h]
    \centering
    \begin{subfigure}{0.72\textwidth}
        \includegraphics[width=\textwidth]{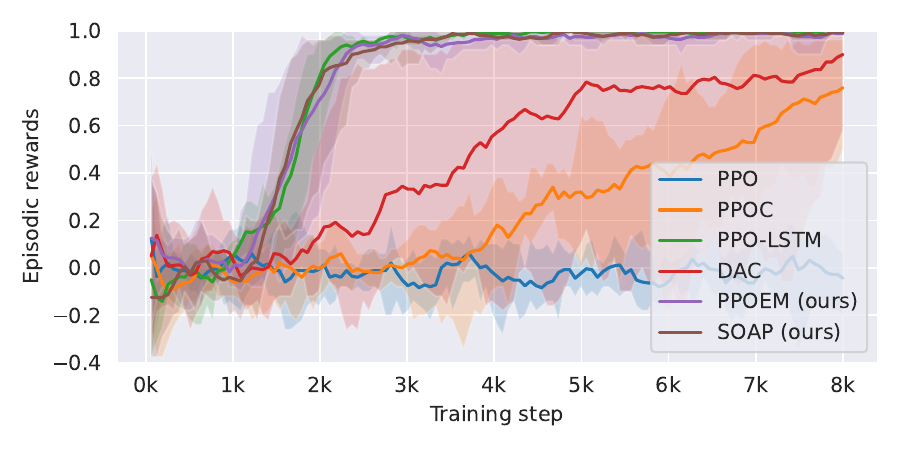}
        \caption{Corridor of length $L=3$}
    \end{subfigure}
    \begin{subfigure}{0.72\textwidth}
        \includegraphics[width=\textwidth]{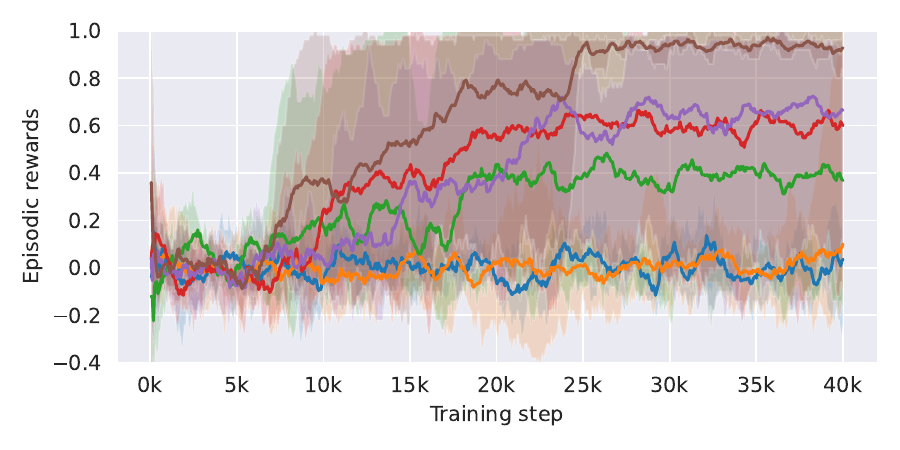}
        \caption{Corridor of length $L=10$}
    \end{subfigure}
    \begin{subfigure}{0.72\textwidth}
        \includegraphics[width=\textwidth]{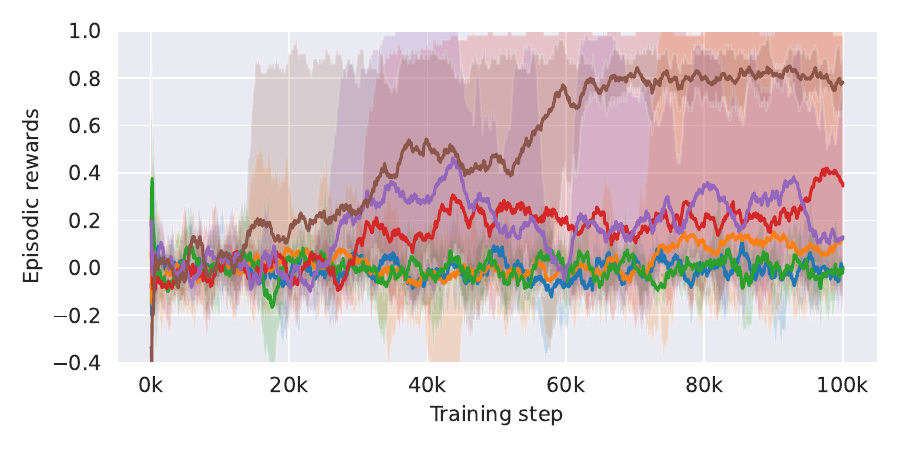}
        \caption{Corridor of length $L=20$}
    \end{subfigure}
    \caption{Training curves of RL agents showing the episodic rewards obtained in the corridor environment with varying lengths. The mean (solid line) and the min-max range (colored shadow) for $5$ seeds per algorithm are shown.}
    \label{fig:ppoem/experiments/corridor_rewards}
\end{figure}

The length of the corridor $L$ can be varied to adjust the difficulty of the task. It is increasingly challenging to retain the information of the starting cell color with longer corridors. In theory, this environment can be solved by techniques such as frame stacking, where the entire history of the agent observations is provided to the policy. However, the computational complexity of this approach scales proportionally to corridor length $L$, which makes this approach unscalable. 

Algorithms with options present an alternative solution, where in theory, the options can be used as latent variables to carry the information relevant to the task. In this experiment, \gls{ppoc}, \gls{ppo_lstm}, \gls{dac}, \gls{ppoem} and \gls{soap} are compared against a standard \gls{ppo} algorithm. The results are shown for corridors with lengths $L=3$, $L=10$ and $L=20$. Due to the increasing level of difficulty of the task, the agents are trained with $8k$, $40k$ and $100k$ time steps of environment interaction, respectively. 

The results are shown in \Cref{fig:ppoem/experiments/corridor_rewards} and \Cref{tab:ppoem/comparison}, and a performance score normalised to the range of the returns of a random agent score and the returns of the best agent is shown in \Cref{tab:ppoem/scores_normalised}. As expected, the vanilla \gls{ppo} agent does not have any memory component so it learned a policy that takes one action deterministically regardless of the color of the first cell. Since the location of the rewarding cell is randomized, this results in an expected return of $0$. 

With \gls{ppoc} that implements the Option-Critic architecture, and \gls{dac} that implements \gls{hrl} using options, while the options should in theory be able to retain information from the past, it could be observed that the training objective was not sufficient to learn a useful option assignment to complete the task. \gls{ppoem} and \gls{soap}, on the other hand, were able to learn to select a different option for a different starting cell color. From \Cref{fig:ppoem/experiments/corridor_rewards}, it could be seen that the two algorithms had identical performance for a short corridor, but as the corridor length $L$ increased, the performance of \gls{ppoem} deteriorated, while \gls{soap} was able to reliably find a correct option assignment, albeit with more training steps. 

There are several major differences between the baseline option-based algorithms (\gls{ppoc} and \gls{dac}) and the proposed algorithms (\gls{ppoem} and \gls{soap}) which could be contributing to their significant differences in performance. Firstly, while the option transition function in \gls{ppoem} and \gls{soap} are in the form of $\pi_\phi(z_{t+1} | s_t, a_t, z_t)$, which allows the assignment of the new option to be conditional on the current option, the option transition in the Option-Critic architecture is decoupled into an option termination probability $\varpi(s_t, z_{t-1})$, and an unconditional inter-option policy $\pi(z_t|s_t)$. This means that whenever the previous option $z_{t-1}$ is terminated with probability $\varpi(s_t, z_{t-1})$, the choice of the new option $z_t$ will be uninformed of the previous option $z_{t-1}$, whereas in \gls{ppoem} and \gls{soap} the probability of the next option $z_{t+1}$ is conditional on the previous option $z_t$. 
The formulation of \gls{dac}~\citep{zhang2019dac} does not have this specific constraint; however, the original implementation by the authors similarly decouples the option transition function such that a new option cannot be fully conditioned on the previous option.

Secondly, both \gls{ppoc} and \gls{dac} rely on learning an option-value function (a value function $V(s, o)$ that is both conditional on the current state $s$ and the current option $o$) to learn the high-level inter-option policy. However, learning an option-value function for an optimal policy can only happen once the inter-option policy is properly learned, but since the inter-option policy is randomly initialized and the option assignment carries little information, it is not possible for the sub-policies to learn optimal policies. 
In the case where the agents' history carries information about the future (e.g. the corridor environment), it is important that the option assignment correctly captures this information without it being lost for the agent to be able to learn an optimal policy. Due to this chicken-and-an-egg problem of learning the inter- and the intra-option policies, neither of these approaches succeeds. In contrast, \gls{soap} directly propagates gradients backwards over multiple timesteps so that the option assignments are directly updated. 

Thirdly, in \gls{ppoc} and \gls{dac}, a new option is sampled at every time step, but the complete option forward distribution given the history is not available as a probability distribution. In contrast, in \gls{ppoem} and \gls{soap} this is available as $\zeta(z_t) \defeq p(z_t | s_{0:t}, a_{0:t-1})$. Evaluating expectations over distributions gives a more robust estimate of the objective function compared to taking a Monte Carlo estimate of the expectations with the sampled options, which is another explanation of why \gls{ppoem} and \gls{soap} were able to learn better option assignments than \gls{ppoc} or \gls{dac}.

\gls{soap}'s training objective maximizes the expectation of returns taken over an option probability conditioned only on the agent's past history, whereas \gls{ppoem}'s objective assumes a fully known trajectory to be able to evaluate the option assignment probability. 
Since option assignments have to be determined online during rollouts, the training objective of \gls{soap} better reflects the task objective. This explains its more reliable performance for longer sequences. 

\gls{ppo_lstm} achieved competitive performance in a corridor with $L=3$, demonstrating the capability of latent states to retain past information, but its performance quickly deteriorated for longer corridors. It could be hypothesized that this is because the latent state space of the recurrent policies is not well constrained, unlike options that take discrete values. Learning a correct value function $V(s, z)$ requires revisiting the same state-latent pair. It is conceivable that with longer sequence lengths during inference time, the latent state will fall within a region that has not been trained well due to compounding noise, leading to an inaccurate estimate of the values and sub-policy. 

\begin{table}[t]
\centering
\caption{Normalized performance comparison of \gls{rl} agents. The agent scores are the returns after the maximum environment steps during training ($100k$ for CartPole, $1M$ for LunarLander and MuJoCo environments, and $10M$ for Atari environments), normalized so that the score of a random agent is 0 and the score of the best performing model is 1. Scores are averaged per environment class (i.e. results for the corridor environments, Atari, and MuJoCo are grouped together) and the \textbf{bold fonts} show the best average normalized score per environment class, while the \textcolor{blue}{blue fonts} show the best-normalized score per environment.}

\begin{tabular}{lrrrrrr}
\toprule
& \multicolumn{2}{c}{Algo w/o options} & \multicolumn{4}{c}{Algo with options}
\\\cmidrule(lr){2-3}\cmidrule(lr){4-7}
Environment         & \multicolumn{1}{c}{PPO}  & \multicolumn{1}{c}{PPO-LSTM} & \multicolumn{1}{c}{PPOC} & \multicolumn{1}{c}{DAC}  & \multicolumn{1}{c}{PPOEM (ours)} & \multicolumn{1}{c}{SOAP (ours)} \\ 
\midrule
\textbf{Corridor} & -0.05 & 0.43 & 0.31 & 0.65 & 0.60 & \textbf{1.00} \\
\quad $L=3$ & -0.08 & 1.00 & 0.76 & 0.90 & 0.99 & \textcolor{blue}{1.00} \\
\quad $L=10$ & -0.01 & 0.36 & 0.06 & 0.63 & 0.70 & \textcolor{blue}{1.00} \\
\quad $L=20$ & -0.05 & -0.06 & 0.11 & 0.41 & 0.12 & \textcolor{blue}{1.00} \\
\midrule
\textbf{CartPole} & \textbf{1.00} & 0.98 & 0.80 & \textbf{1.00} & 0.98 & \textbf{1.00} \\
\midrule
\textbf{LunarLander} & 0.86 & \textbf{1.00} & 0.74 & 0.78 & 0.99 & 0.99 \\
\midrule
\textbf{Atari} & \textbf{0.93} & 0.78 & 0.22 & 0.85 & 0.74 & 0.89 \\
\quad Asteroids & 0.83 & 0.64 & 0.81 & \textcolor{blue}{1.00} & 0.93 & 0.89 \\
\quad Beam Rider & 0.93 & 0.37 & 0.13 & 0.70 & 0.66 & \textcolor{blue}{1.00} \\
\quad Breakout & \textcolor{blue}{1.00} & 0.68 & 0.01 & 0.95 & 0.14 & 0.92 \\
\quad Enduro & 0.97 & 0.90 & 0.00 & 0.93 & \textcolor{blue}{1.00} & 0.82 \\
\quad Ms Pacman & 0.88 & 0.74 & 0.15 & 0.87 & 0.69 & \textcolor{blue}{1.00} \\
\quad Pong & 1.00 & \textcolor{blue}{1.00} & 0.48 & \textcolor{blue}{1.00} & 0.94 & \textcolor{blue}{1.00} \\
\quad Qbert & \textcolor{blue}{1.00} & 0.97 & 0.00 & 0.65 & 0.70 & 0.90 \\
\quad Road Runner & \textcolor{blue}{1.00} & 0.92 & 0.15 & 0.88 & 0.52 & 0.81 \\
\quad Seaquest & 0.89 & 0.87 & 0.18 & 0.69 & \textcolor{blue}{1.00} & 0.55 \\
\quad Space Invaders & 0.82 & 0.73 & 0.32 & 0.82 & 0.82 & \textcolor{blue}{1.00} \\
\midrule
\textbf{MuJoCo} & \textbf{0.97} & 0.75 & 0.60 & 0.82 & 0.43 & 0.93 \\
\quad Ant & \textcolor{blue}{1.00} & 0.46 & 0.07 & 0.64 & 0.07 & 0.87 \\
\quad Half Cheetah & \textcolor{blue}{1.00} & 0.83 & 0.80 & 0.79 & 0.01 & 0.92 \\
\quad Humanoid & 0.98 & 0.96 & 0.96 & \textcolor{blue}{1.00} & 0.13 & 0.90 \\
\quad Reacher & 0.99 & 0.99 & 0.99 & \textcolor{blue}{1.00} & 0.98 & \textcolor{blue}{1.00} \\
\quad Swimmer & 0.96 & 0.97 & 0.34 & \textcolor{blue}{1.00} & 0.95 & 0.90 \\
\quad Walker & 0.85 & 0.31 & 0.47 & 0.52 & 0.44 & \textcolor{blue}{1.00} \\
\bottomrule
\end{tabular}

\label{tab:ppoem/scores_normalised}
\end{table}

\subsection{Stability of the algorithms on CartPole, LunarLander, Atari, and MuJoCo environments}

Experiments were also conducted on standard \gls{rl} environments to evaluate the stability of the algorithms with options. Results for CartPole-v1 and LunarLander-v2 are shown in \Cref{fig:ppoem/experiments/cartpole_lunarlander}, and results on 10 Atari environments~\citep{bellemare13arcade} and 6 MuJoCo environments~\citep{todorov2012mujoco} are shown in \Cref{fig:ppoem/experiments/atari} and \Cref{fig:ppoem/experiments/mujoco}, respectively. \Cref{tab:ppoem/comparison} summarises the agent scores after the maximum environment steps during training ($100k$ for CartPole, $1M$ for LunarLander and MuJoCo environments, and $10M$ for Atari environments), and \Cref{tab:ppoem/scores_normalised} shows the scores normalized so that the score of a random agent is 0 and the score of the best performing model is 1. (If an agent's final score is lower than a random agent it can have negative normalized scores.) 
There was no significant difference in performances amongst the algorithms for simpler environments such as CartPole and LunarLander, with \gls{ppoc} and \gls{dac} having slightly worse performance than others. For the Atari and MuJoCo environments, however, there was a consistent trend that \gls{soap} achieves similar performances (slightly better in some cases, slightly worse in others) to the vanilla \gls{ppo}, while \gls{ppoem}, \gls{ppo_lstm} and \gls{ppoc} were significantly less stable to train. It could be hypothesized that, similarly to \Cref{sec:ppoem/experiments/corridor}, the policy of \gls{ppoc} disregarded the information of past options when choosing the next option, which is why the performance was unstable with larger environments. Another point of consideration is that, with $N$ number of options, there are $N$ number of sub-policies to train, which becomes increasingly computationally expensive and requires many visits to the state-option pair in the training data, especially when using a Monte Carlo estimate by sampling the next option as is done in \gls{ppoc} and \gls{dac} instead of maintaining a distribution of the option $\zeta(z_t)$ as in \gls{ppoem} and \gls{soap}. As for \gls{ppo_lstm}, similar reasoning as in \Cref{sec:ppoem/experiments/corridor} suggests that with complex environments with a variety of trajectories that can be taken through the state space, the latent states that could be visited increases combinatorially, making it challenging to learn a robust sub-policy and value functions. 

\section{Conclusion}
Two competing algorithms, \gls{ppoem} and \gls{soap}, are proposed to solve the problem of option discovery and assignments in an unsupervised way. \gls{ppoem} implements a training objective of maximizing the expected returns using the \gls{em} algorithm, while \gls{soap} analytically evaluates the policy gradient of the option policy to derive an option advantage function that facilitates temporal propagation of the policy gradients.
These approaches have an advantage over Option-Critic architecture in that (a) the option distribution is analytically evaluated rather than sampled, and (b) the option transitions are fully conditional on the previous option, allowing historical information to propagate forward in time beyond the temporal window provided as observations.

Experiments in \gls{pomdp} corridor environments designed to require options showed that \gls{soap} is the most robust way of learning option assignments that adhere to the task objective. \gls{soap} also maintained its performance when solving \gls{mdp} tasks without the need for options (e.g. Atari with frame-stacking), whereas \gls{ppoc}, \gls{dac}, \gls{ppo_lstm} and \gls{ppoem} were less stable when solving these problems.

\section{Future Work}
\gls{soap} demonstrated capabilities of learning options in a \gls{pomdp} environment of corridors, and showed equivalent performances to the baseline \gls{ppo} agent in other environments. However, even in simple settings, it took the agent many samples before a correct option assignment was learned. Option discovery is a difficult chicken and an egg problem, since options need to be assigned correctly in order for the rewards to be obtained and passed onto the options, but without the rewards a correct option assignment may not be learned. Furthermore, learning to segment episodes into options in an unsupervised way without any pre-training is an ill-defined problem, since there could be many equally valid solutions. Combining the learning objective of \gls{soap} with methods such as curriculum learning to pre-train diverse sub-policies specialized to different tasks may stabilize training.

In the current formulation of \gls{soap}, the options are discrete variables, and are less expressive compared to latent variables in recurrent policies and transformers. This greatly reduces the memory capacity and could hinder learning in \gls{pomdp} environments. Further research in extending the derivations of \gls{soap} to work with continuous or multi-discrete variables as latents may lead to making the method scalable to more complex problems.

\section*{Acknowledgements} 
The authors acknowledge the generous support of the Royal Academy of Engineering (RF\textbackslash 201819\textbackslash 18\textbackslash 163) and the Ezoe Memorial Recruit Foundation.
For the purpose of Open Access, the authors have applied a CC BY public copyright licence to any Author Accepted Manuscript (AAM) version arising from this submission (\url{https://openaccess.ox.ac.uk/rights-retention/}).

\bibliography{main}
\bibliographystyle{tmlr}

\newpage

\appendix
\section{Appendix}

\begin{table}[h]
\centering
\caption{The returns of \gls{rl} agents after the maximum environment training steps ($100k$ for CartPole, $1M$ for LunarLander and MuJoCo, and $10M$ for Atari).}
\label{tab:ppoem/comparison}\begin{tabular}{lrrrrrrr}
\toprule
& \multicolumn{2}{c}{Algo w/o options} & \multicolumn{4}{c}{Algo with options}
\\\cmidrule(lr){2-3}\cmidrule(lr){4-7}
Environment         & \multicolumn{1}{c}{PPO}  & \multicolumn{1}{c}{PPO-LSTM} & \multicolumn{1}{c}{PPOC} & \multicolumn{1}{c}{DAC}  & \multicolumn{1}{c}{PPOEM (ours)} & \multicolumn{1}{c}{SOAP (ours)} \\ 
\midrule
Corridor $L=3$         & -0.04         & 0.99          & 0.76              & 0.90          & 0.99                  & 0.99                 \\ 
Corridor $L=10$        & 0.04          & 0.37          & 0.10              & 0.60          & 0.66                  & 0.93                 \\ 
Corridor $L=20$        & 0.00          & 0.00          & 0.13              & 0.34          & 0.13                  & 0.78                 \\ 
CartPole             & 499.65        & 492.75        & 401.87            & 500.00        & 491.30                & 500.00               \\ 
LunarLander          & 195.03        & 257.96        & 141.07            & 159.00        & 251.71                & 254.14               \\ 
Asteroids            & 1806.14       & 1463.03       & 1760.53           & 2108.87       & 1985.70               & 1910.53              \\ 
Beam Rider            & 3704.50       & 1574.99       & 689.07            & 2839.36       & 2691.09               & 3955.31              \\ 
Breakout             & 376.19        & 257.57        & 4.64              & 355.97        & 51.02                 & 345.68               \\ 
Enduro                & 704.17        & 655.96        & 0.13              & 674.03        & 724.97                & 593.44               \\ 
Ms Pacman               & 1825.40       & 1555.80       & 434.17            & 1806.40       & 1453.87               & 2046.87              \\ 
Pong                 & 20.74         & 20.70         & -0.99             & 20.65         & 18.32                 & 20.81                \\ 
Qbert                & 12669.59      & 12282.50      & 115.17            & 8212.08       & 8940.83               & 11469.25             \\ 
Road Runner           & 36247.13      & 33507.00      & 5273.33           & 32042.00      & 18886.00              & 29490.33             \\ 
Seaquest             & 1539.73       & 1501.13       & 317.80            & 1198.33       & 1723.50               & 951.67               \\ 
Space Invaders        & 845.73        & 756.27        & 378.90            & 840.63        & 844.00                & 1013.40              \\ 
Ant                  & 2258.28       & 987.01        & 53.10             & 1394.61       & 49.30                 & 1943.45              \\ 
Half Cheetah          & 5398.50       & 4478.37       & 4300.07           & 4251.44       & 56.07                 & 4962.48              \\ 
Humanoid             & 1196.32       & 1169.12       & 1169.26           & 1212.80       & 275.03                & 1100.38              \\ 
Reacher              & -4.87         & -5.04         & -4.94             & -4.60         & -5.54                 & -4.87                \\ 
Swimmer              & 340.31        & 342.84        & 119.46            & 354.67        & 337.85                & 319.45               \\ 
Walker               & 2960.18       & 1071.81       & 1631.97           & 1799.50       & 1517.90               & 3478.50              \\ 
\bottomrule
\end{tabular}
\end{table}

\begin{figure}[h]
    \centering
    \begin{subfigure}{0.48\textwidth}
        \includegraphics[width=1.0\textwidth]{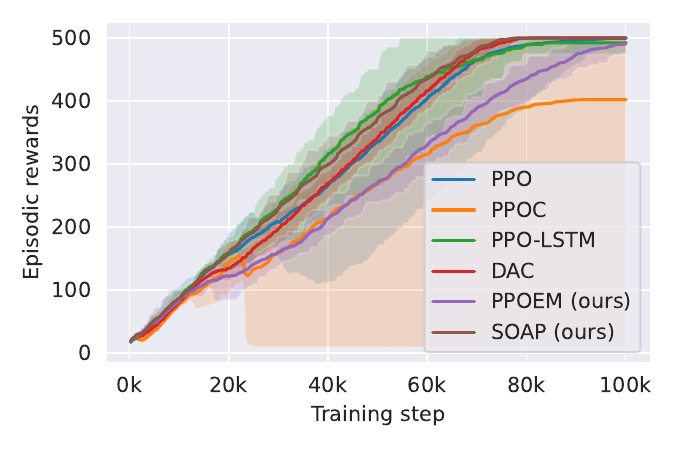}
        \caption{CartPole-v1}
    \end{subfigure}
    \begin{subfigure}{0.48\textwidth}
        \includegraphics[width=1.0\textwidth]{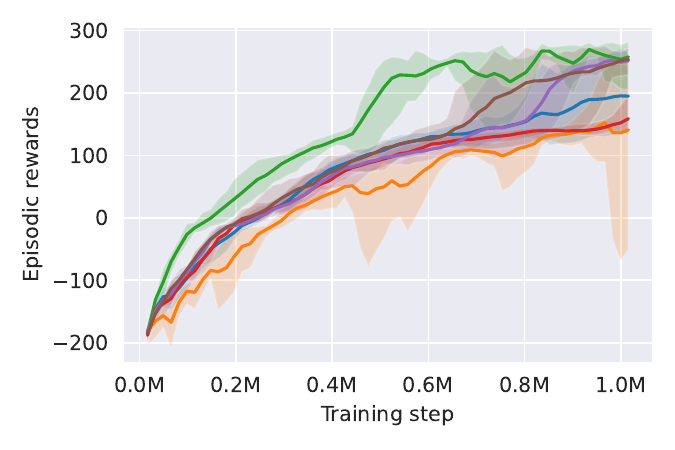}
        \caption{LunarLander-v2}
    \end{subfigure}
    \caption{Training curves of RL agents showing the episodic rewards obtained in the CartPole-v1 and LunarLander-v2 environments. The mean (solid line) and the min-max range (colored shadow) for $5$ seeds per algorithm are shown.}
    \label{fig:ppoem/experiments/cartpole_lunarlander}
\end{figure}

\begin{figure}[h]
    \centering
    \begin{subfigure}{0.48\textwidth}
        \includegraphics[width=1.0\textwidth]{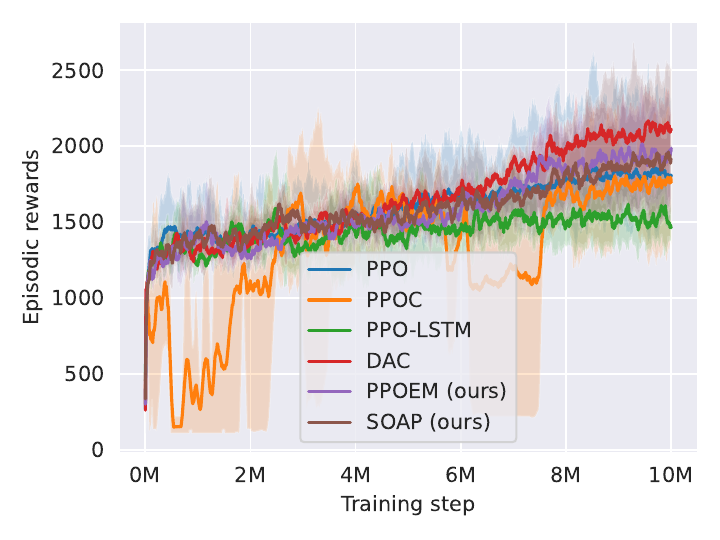}
        \caption{Astroids}
    \end{subfigure}
    \begin{subfigure}{0.48\textwidth}
        \includegraphics[width=1.0\textwidth]{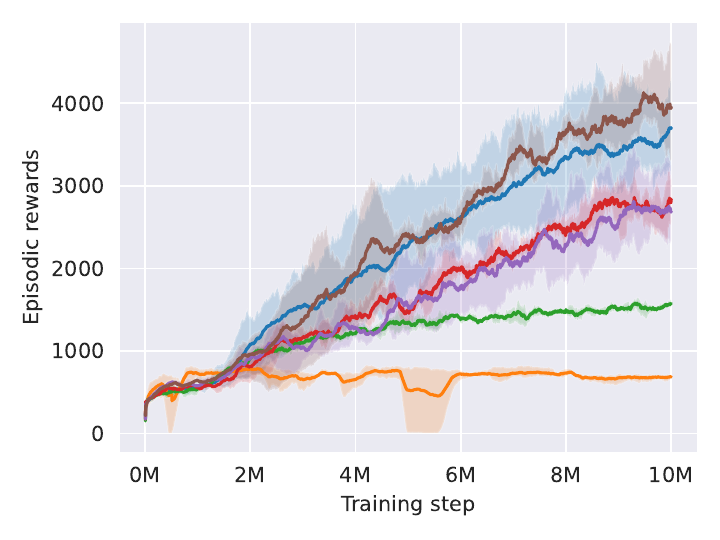}
        \caption{Beam Rider}
    \end{subfigure}
    \begin{subfigure}{0.48\textwidth}
        \includegraphics[width=1.0\textwidth]{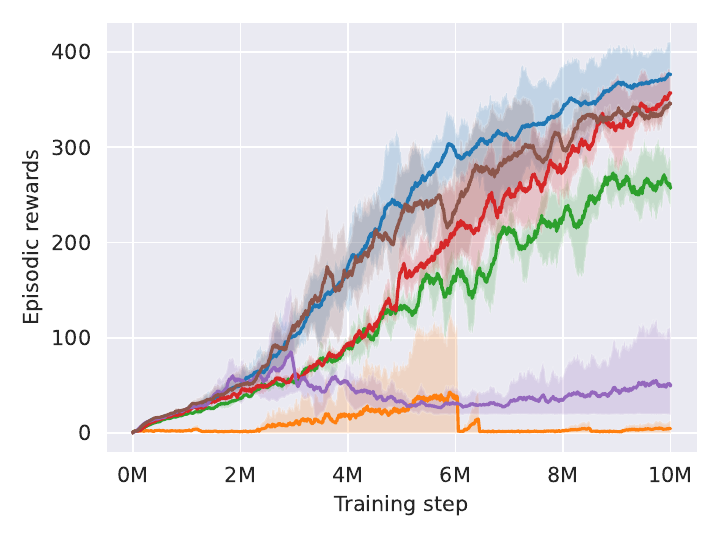}
        \caption{Breakout}
    \end{subfigure}
    \begin{subfigure}{0.48\textwidth}
        \includegraphics[width=1.0\textwidth]{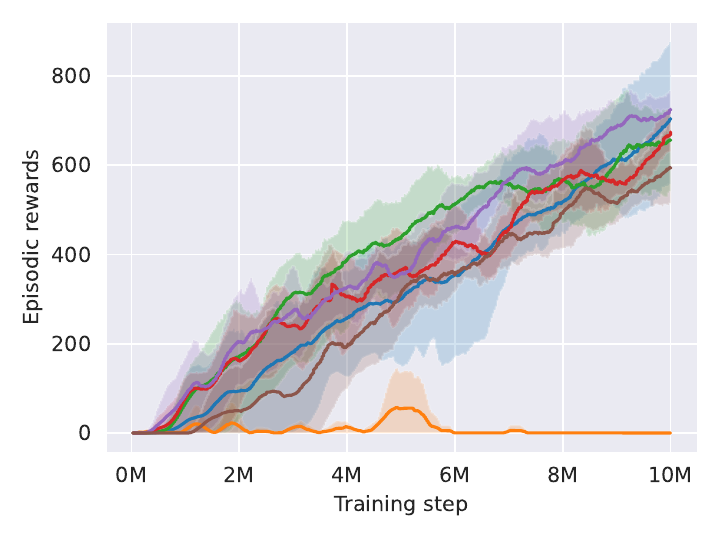}
        \caption{Enduro}
    \end{subfigure}
    \caption{Training curves of RL agents showing the episodic rewards obtained in the Atari environments. The mean (solid line) and the min-max range (colored shadow) for $3$ seeds per algorithm are shown. [Spans multiple pages]}
    \label{fig:ppoem/experiments/atari}
\end{figure}

\begin{figure}[h]
    \ContinuedFloat
    \setcounter{subfigure}{4}
    \centering
    \begin{subfigure}{0.48\textwidth}
        \includegraphics[width=\textwidth]{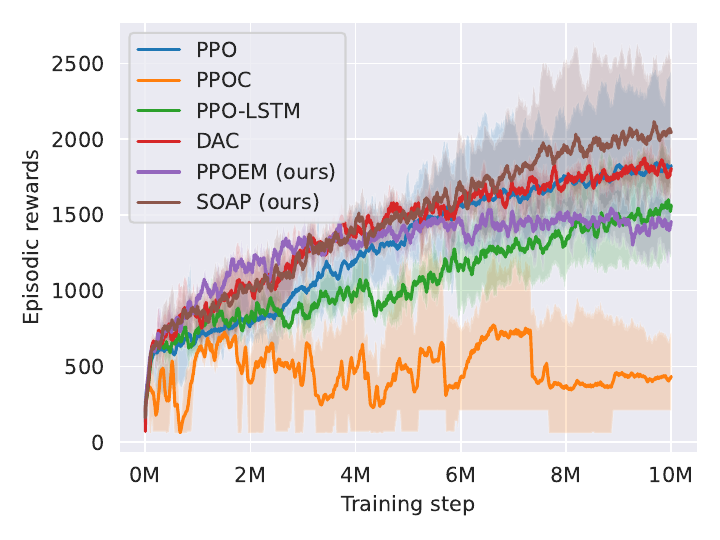}
        \caption{Ms Pacman}
    \end{subfigure}
    \begin{subfigure}{0.48\textwidth}
        \includegraphics[width=\textwidth]{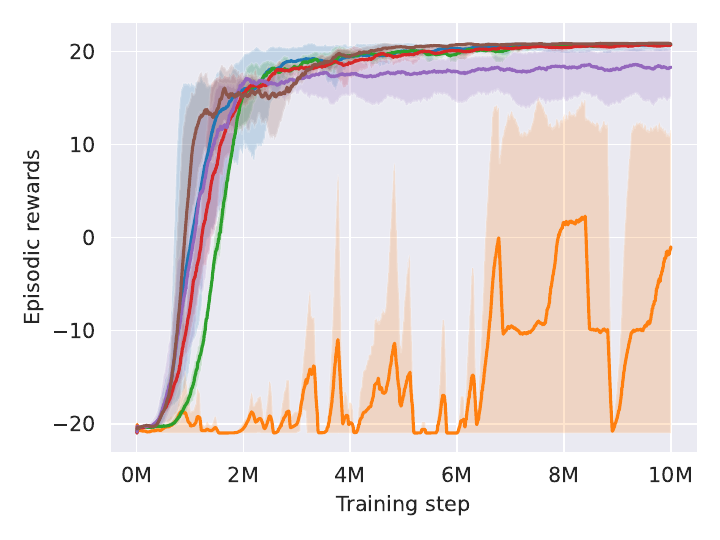}
        \caption{Pong}
    \end{subfigure}
    \begin{subfigure}{0.48\textwidth}
        \includegraphics[width=\textwidth]{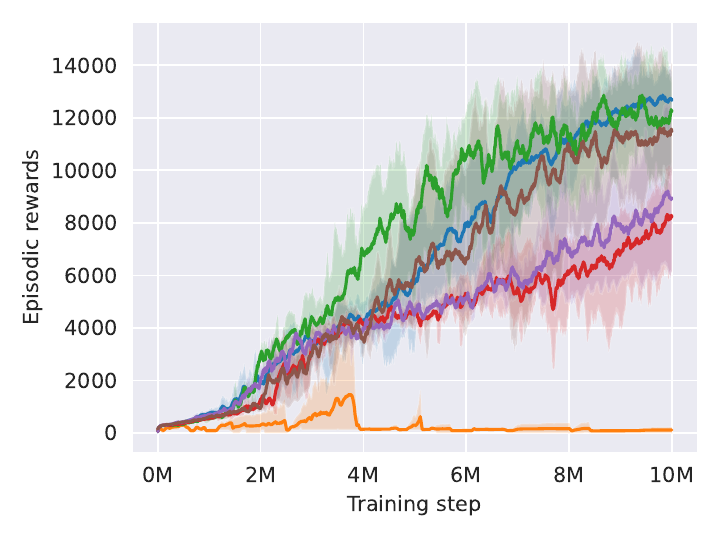}
        \caption{Qbert}
    \end{subfigure}
    \begin{subfigure}{0.48\textwidth}
        \includegraphics[width=\textwidth]{figures/ppoem/plots/beamrider.pdf}
        \caption{Road Runner}
    \end{subfigure}
    \begin{subfigure}{0.48\textwidth}
        \includegraphics[width=\textwidth]{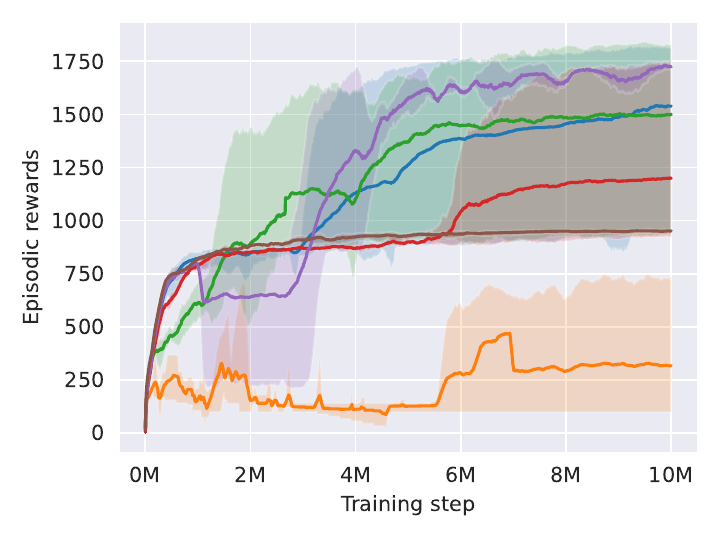}
        \caption{Seaquest}
    \end{subfigure}
    \begin{subfigure}{0.48\textwidth}
        \includegraphics[width=\textwidth]{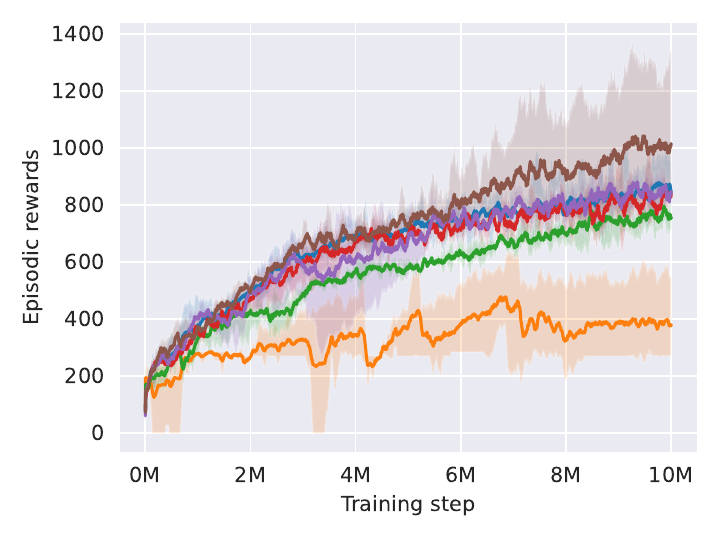}
        \caption{Space Invader}
    \end{subfigure}
    \caption*{[Continued] Training curves of RL agents showing the episodic rewards obtained in the Atari environments. The mean (solid line) and the min-max range (colored shadow) for $3$ seeds per algorithm are shown.}
\end{figure}

\begin{figure}[h]
    \centering
    \begin{subfigure}{0.48\textwidth}
        \includegraphics[width=\textwidth]{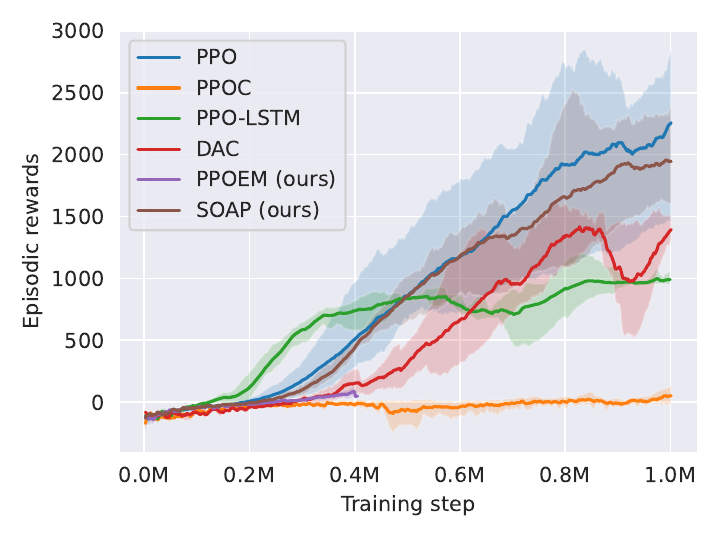}
        \caption{Ant}
    \end{subfigure}
    \begin{subfigure}{0.48\textwidth}
        \includegraphics[width=\textwidth]{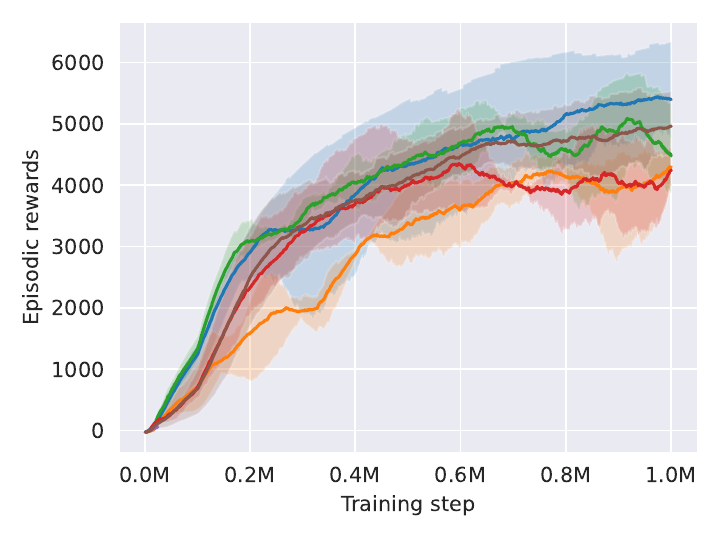}
        \caption{HalfCheetah}
    \end{subfigure}
    \begin{subfigure}{0.48\textwidth}
        \includegraphics[width=\textwidth]{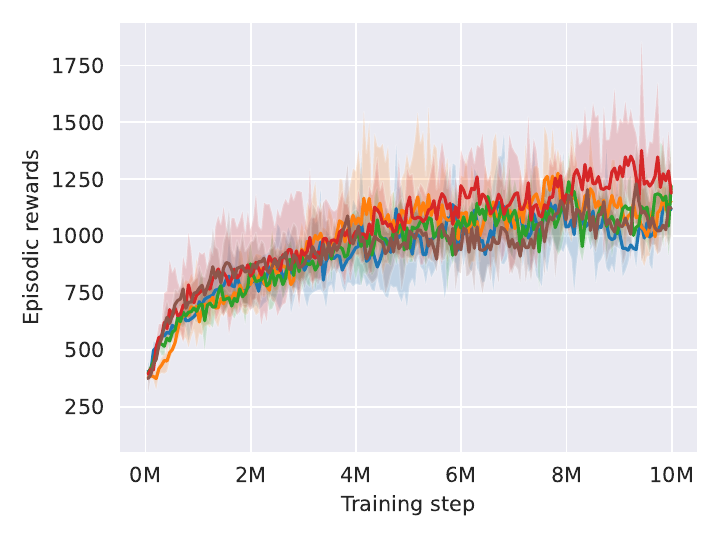}
        \caption{Humanoid}
    \end{subfigure}
    \begin{subfigure}{0.48\textwidth}
        \includegraphics[width=\textwidth]{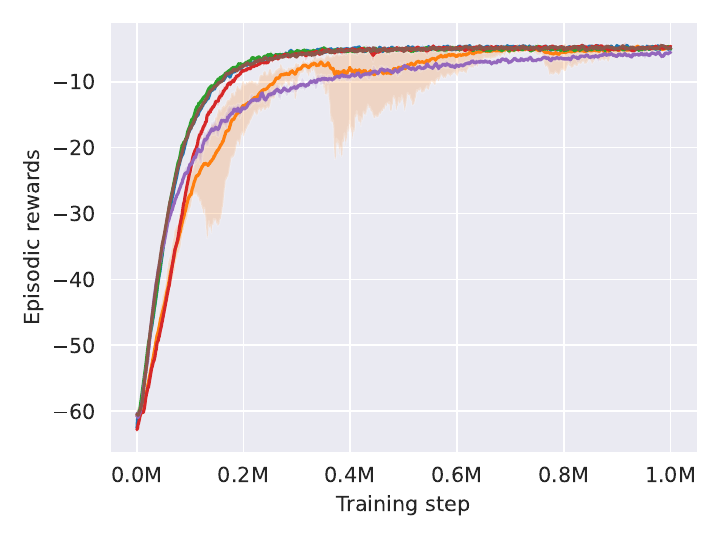}
        \caption{Reacher}
    \end{subfigure}
    \begin{subfigure}{0.48\textwidth}
        \includegraphics[width=\textwidth]{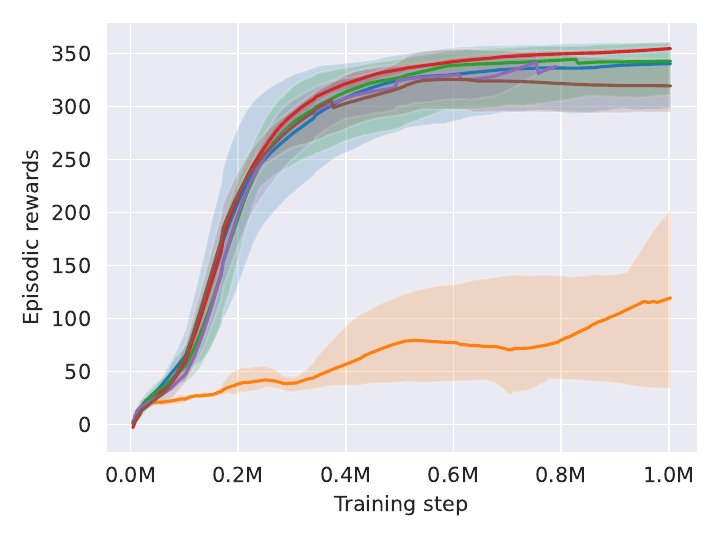}
        \caption{Swimmer}
    \end{subfigure}
    \begin{subfigure}{0.48\textwidth}
        \includegraphics[width=\textwidth]{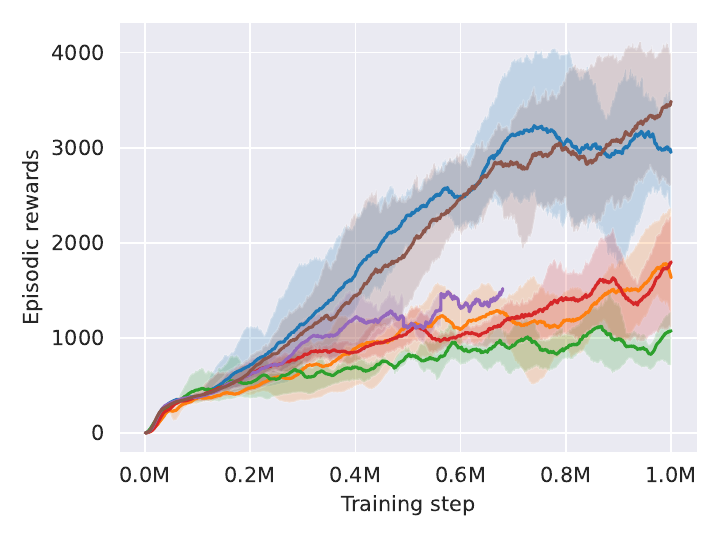}
        \caption{Walker}
    \end{subfigure}
    \caption{Training curves of RL agents showing the episodic rewards obtained in the MuJoCo environments. The mean (solid line) and the min-max range (colored shadow) for $3$ seeds per algorithm are shown. Note that the PPOEM algorithm failed mid-way in some cases due to training instabilities.}
    \label{fig:ppoem/experiments/mujoco}
\end{figure}

\end{document}

%% file: math_commands.tex
\usepackage{amsmath,amsfonts,bm}

\def\eqref#1{equation~\ref{#1}}

\def\1{\bm{1}}

\DeclareMathAlphabet{\mathsfit}{\encodingdefault}{\sfdefault}{m}{sl}
\SetMathAlphabet{\mathsfit}{bold}{\encodingdefault}{\sfdefault}{bx}{n}

\DeclareMathOperator*{\argmax}{arg\,max}

%% file: glossary.tex
\newglossaryentry{mdp}{
    name={MDP},
    description={Markov Decision Process},
    first={Markov Decision Process (MDP)},
    firstplural={Markov Decision Processes (MDPs)},
    plural=MDPs
}

\newglossaryentry{pomdp}{
    name={POMDP},
    description={Partially Observable Markov Decision Process},
    first={Partially Observable Markov Decision Process (POMDP)},
    firstplural={Partially Observable Markov Decision Processes (POMDPs)},
    plural=POMDPs
}

\newglossaryentry{il}{
    name={IL},
    description={Imitation Learning},
    first={Imitation Learning (IL)}
}

\newglossaryentry{rl}{
    name={RL},
    description={Reinforcement Learning},
    first={Reinforcement Learning (RL)}
}

\newglossaryentry{bc}{
    name={BC},
    description={Behavioural Cloning},
    first={Behavioural Cloning (BC)}
}

\newglossaryentry{irl}{
    name={IRL},
    description={Inverse Reinforcement Learning},
    first={Inverse Reinforcement Learning (IRL)}
}

\newglossaryentry{hrl}{
    name={HRL},
    description={Hierarchical Reinforcement Learning},
    first={Hierarchical Reinforcement Learning (HRL)}
}

\newglossaryentry{gail}{
    name={GAIL},
    description={Generative Adversarial Imitation Learning},
    first={Generative Adversarial Imitation Learning (GAIL)}
}

\newglossaryentry{meta-rl}{
    name={Meta-RL},
    description={Meta Reinforcement Learning},
    first={Meta Reinforcement Learning (Meta-RL)}
}

\newglossaryentry{vin}{
    name={VIN},
    description={Value Iteration Network},
    first={Value Iteration Network (VIN)},
    firstplural={Value Iteration Networks (VINs)},
    plural=VINs
}

\newglossaryentry{llm}{
    name={LLM},
    description={Large Language Model},
    first={Large Language Model (LLM)},
    firstplural={Large Language Models (LLMs)},
    plural=LLMs
}

\newglossaryentry{em}{
    name={EM},
    description={Expectation Maximisation},
    first={Expectation Maximisation (EM)},
}

\newglossaryentry{hmm}{
    name={HMM},
    description={Hidden Markov Model},
    first={Hidden Markov Model (HMM)},
    firstplural={Hidden Markov Model (HMMs)},
    plural=HMMs
}

\newglossaryentry{ppo}{
    name={PPO},
    description={Proximal Policy Optimisation},
    first={Proximal Policy Optimisation (PPO)},
}

\newglossaryentry{calvin}{
    name={CALVIN},
    description={Collision Avoidance Long-term Value Iteration Network},
    first={Collision Avoidance Long-term Value Iteration Network (CALVIN)},
}

\newglossaryentry{lstm}{
    name={LSTM},
    description={Long Short-Term Memory},
    first={Long Short-Term Memory (LSTM)},
    plural=LSTMs
}

\newglossaryentry{gae}{
    name={GAE},
    description={Generalised Advantage Estimate},
    first={Generalised Advantage Estimate (GAE)}
}

\newglossaryentry{goa}{
    name={GOA},
    description={Generalised Option Advantage},
    first={Generalised Option Advantage (GOA)}
}

\newglossaryentry{td}{
    name={TD},
    description={Temporal Difference},
    first={Temporal Difference (TD)}
}

\newglossaryentry{gru}{
    name={GRU},
    description={Gated Recurrent Unit},
    first={Gated Recurrent Unit (GRU)},
    firstplural={Gated Recurrent Units (GRUs)},
    plural=GRUs
}

\newglossaryentry{rnn}{
    name={RNN},
    description={Recurrent Neural Network},
    first={Recurrent Neural Network (RNN)},
    firstplural={Recurrent Neural Networks (RNNs)},
    plural=RNNs
}

\newglossaryentry{cnn}{
    name={CNN},
    description={Convolutional Neural Network},
    first={Convolutional Neural Network (CNN)},
    firstplural={Convolutional Neural Networks (CNNs)},
    plural=CNNs
}

\newglossaryentry{gpu}{
    name={GPU},
    description={Graphical Processing Unit},
    first={Graphical Processing Unit (GPU)},
    firstplural={Graphical Processing Units (GPUs)},
    plural=GPUs
}

\newglossaryentry{ppoem}{
    name={PPOEM},
    description={Proximal Policy Optimisation via Expectation Maximisation},
    first={Proximal Policy Optimisation via Expectation Maximisation (PPOEM)},
}

\newglossaryentry{soap}{
    name={SOAP},
    description={Sequential Option Advantage Propagation},
    first={Sequential Option Advantage Propagation (SOAP)},
}

\newglossaryentry{ppoc}{
    name={PPOC},
    description={Proximal Policy Option-Critic},
    first={Proximal Policy Option-Critic (PPOC)},
}

\newglossaryentry{ale}{
    name={ALE},
    description={Arcade Learning Environment},
    first={Arcade Learning Environment (ALE)},
}

\newglossaryentry{ppo_lstm}{
    name={PPO-LSTM},
    description={Proximal Policy Optimisation with Long Short-Term Memory},
    first={Proximal Policy Optimisation with Long Short-Term Memory (PPO-LSTM)},
}

\newglossaryentry{ai}{
    name={AI},
    description={Artificial Intelligence},
    first={Artificial Intelligence (AI)},
}

\newglossaryentry{vi}{
    name={VI},
    description={Value Iteration},
    first={Value Iteration (VI)},
}

\newglossaryentry{mcts}{
    name={MCTS},
    description={Monte Carlo Tree Search},
    first={Monte Carlo Tree Search (MCTS)}
}

\newglossaryentry{idm}{
    name={IDM},
    description={Inverse Dynamics Model},
    first={Inverse Dynamics Model (IDM)},
}

\newglossaryentry{vpt}{
    name={VPT},
    description={Video PreTraining},
    first={Video PreTraining (VPT)},
}

\newglossaryentry{prm}{
    name={PRM},
    description={Probabilistic Roadmap},
    first={Probabilistic Roadmap (PRM)},
    firstplural={Probabilistic Roadmaps (PRM)},
    plural={PRM}
}

\newglossaryentry{rrt}{
    name={RRT},
    description={Probabilistic Roadmap},
    first={Rapidly-exploring Random Tree (RRT)},
    firstplural={Rapidly-exploring Random Trees (RRT)},
    plural={RRT}
}

\newglossaryentry{gan}{
    name={GAN},
    description={Generative Adversarial Network},
    first={Generative Adversarial Network (GAN)},
    firstplural={Generative Adversarial Networks (GANs)},
    plural={GANs}
}

\newglossaryentry{a2c}{
    name={A2C},
    description={Advantage Actor-Critic},
    first={Advantage Actor-Critic (A2C)},
}

\newglossaryentry{trpo}{
    name={TRPO},
    description={Trust Region Policy Optimisation},
    first={Trust Region Policy Optimisation (TRPO)},
}

\newglossaryentry{kl}{
    name={KL},
    description={Kullback–Leibler},
    first={Kullback–Leibler (KL)},
}

\newglossaryentry{ddpg}{
    name={DDPG},
    description={Deep Deterministic Policy Gradients},
    first={Deep Deterministic Policy Gradients (DDPG)},
}

\newglossaryentry{td3}{
    name={TD3},
    description={Twin-Delayed Deep Deterministic Policy Gradients},
    first={Twin-Delayed DDPG (TD3)},
}

\newglossaryentry{sac}{
    name={SAC},
    description={Soft Actor-Critic},
    first={Soft Actor-Critic (SAC)},
}

\newglossaryentry{vae}{
    name={VAE},
    description={Variational Auto-Encoder},
    first={Variational Auto-Encoder (VAE)},
    firstplural={Variational Auto-Encoders (VAEs)},
    plural={VAEs}
}

\newglossaryentry{slam}{
    name={SLAM},
    description={Simultaneous Localisation and Mapping},
    first={Simultaneous Localisation and Mapping (SLAM)},
}

\newglossaryentry{vlm}{
    name={VLM},
    description={Vision-Language Model},
    first={Vision-Language Model (VLM)},
}

\newglossaryentry{avd}{
    name={AVD},
    description={Active Vision Dataset},
    first={Active Vision Dataset (AVD)},
}

\newglossaryentry{cmp}{
    name={CMP},
    description={Cognitive Mapper and Planner},
    first={Cognitive Mapper and Planner (CMP)},
}

\newglossaryentry{gppn}{
    name={GPPN},
    description={Gated Path Planning Network},
    first={Gated Path Planning Network (GPPN)},
    firstplural={Gated Path Planning Networks (GPPNs)},
    plural={GPPNs}
}

\newglossaryentry{lpn}{
    name={LPN},
    description={Lattice PointNet},
    first={Lattice PointNet (LPN)},
}

\newglossaryentry{relu}{
    name={ReLU},
    description={Rectified Linear Unit},
    first={Rectified Linear Unit (ReLU)},
}

\newglossaryentry{gpt}{
    name={GPT},
    description={Generative Pre-trained Transformer},
    first={Generative Pre-trained Transformer (GPT)},
}

\newglossaryentry{ucb}{
    name={UCB},
    description={Upper Confidence Bound},
    first={Upper Confidence Bound (UCB)},
}

\newglossaryentry{dqn}{
    name={DQN},
    description={Deep Q-Network},
    first={Deep Q-Network (DQN)},
}

\newglossaryentry{ppg}{
    name={PPG},
    description={Phasic Policy Gradient},
    first={Phasic Policy Gradient (PPG)},
}

\newglossaryentry{iic}{
    name={IIC},
    description={Invariant Information Clustering},
    first={Invariant Information Clustering (IIC)},
}

\newglossaryentry{awm}{
    name={AWM},
    description={Abstract World Model},
    first={Abstract World Model (AWM)},
}

\newglossaryentry{dag}{
    name={DAG},
    description={Directed Acyclic Graph},
    first={Directed Acyclic Graph (DAG)},
}

\newglossaryentry{dac}{
    name={DAC},
    description={Double Actor-Critic},
    first={Double Actor-Critic (DAC)},
}

\newglossaryentry{smdp}{
    name={Semi-MDP},
    description={Semi-Markov Decision Process},
    first={Semi-Markov Decision Process (Semi-MDP)},
    firstplural={Semi-Markov Decision Processes (MDPs)},
    plural=Semi-MDPs    
}

\newglossaryentry{ml}{
    name={ML},
    description={Machine Learning},
    first={Machine Learning (ML)},
}